\def\longversion{}\documentclass{IOS-Book-Article}
\usepackage{mathptmx}
\usepackage{soul}\setuldepth{article}
\usepackage{graphicx}
\usepackage{tikz}
\usepackage{tikzscale}
\usetikzlibrary{positioning, shapes.multipart, arrows.meta, backgrounds, fit, quotes}

\usepackage{orcidlink}

\definecolor{Class}{RGB}{207,165,0}      
\definecolor{ObjPro}{RGB}{0,121,186}      

\usepackage{amsmath}
\usepackage{amsfonts}
\usepackage{todonotes}
\usepackage{xspace}
\usepackage{newfloat} 
\usepackage{cleveref}
\usepackage{ntheorem}
\usepackage{stmaryrd}
\usepackage{multirow}
\usepackage{relsize}
\usepackage{booktabs}
\usepackage{ifthen}

\usepackage[edges]{forest}

\usepackage{placeins}

\theoremstyle{numberbreak}
\theorembodyfont{\normalfont}

\theoremstyle{break}
\theorembodyfont{\normalfont}

\newcommand{\riskman}{\textsc{Riskman}\xspace}

\newcommand{\ebnfeq}{\mathrel{::=}}
\newcommand{\ebnfalt}{\mathbin{\mid}}

\newcommand{\define}[1]{\textit{#1}}

\newcommand{\set}[1]{\left\{#1\right\}}
\newcommand{\tuple}[1]{\left(#1\right)}
\newcommand{\guard}{\ \middle\vert\ }
\newcommand{\card}[1]{\left\lvert#1\right\rvert}

\newcommand{\refreq}[2]{} % \refreq{} is the new \eats{}

\newcommand{\vdespec}{VDE Spec 90025\xspace}
\newcommand{\vdespeccite}{\vdespec~\cite{VDESpec24}\xspace}
\newcommand{\vdespecciteo}[1][]{\vdespec~\cite[#1]{VDESpec24}\xspace}
\newcommand{\isonorm}{ISO 14971\xspace}
\newcommand{\isonormcite}{\isonorm~\cite{ISO14971}\xspace}
\newcommand{\eqdef}{\mathrel{:=}}

\tikzstyle{myborder}=[densely dashed]

\newcommand{\probmag}[2]{\raisebox{0pt}[1ex][0pt]{#1:\,#2}}
\newcommand{\represents}{\mathbin{\hat{=}}}

\newcommand{\Pone}{\ensuremath{\mathit{P1}}\xspace}
\newcommand{\Ptwo}{\ensuremath{\mathit{P2}}\xspace}
\newcommand{\riskmanurl}[1][]{\url{https://w3id.org/riskman#1}\xspace}
\newcommand{\ontologyurl}[1][]{\riskmanurl[/ontology]}
\newcommand{\shapesurl}[1][]{\riskmanurl[/shapes]}
\newcommand{\dlfont}[1]{\ensuremath{\mathsf{#1}}}

\newcommand{\grt}{\ensuremath{\dlfont{gt}}}
\newcommand\sbullet[1][.5]{\mathbin{\vcenter{\hbox{\scalebox{#1}{$\bullet$}}}}}

\newcommand{\ARisk}{\dlfont{AnalyzedRisk}}
\newcommand{\ASDA}{\dlfont{AssuranceSDA}}
\newcommand{\ASDAI}{\dlfont{AssuranceSDAI}}
\newcommand{\CRisk}{\dlfont{ControlledRisk}}
\newcommand{\DComponent}{\dlfont{DeviceComponent}}
\newcommand{\DContext}{\dlfont{DeviceContext}}
\newcommand{\DFunction}{\dlfont{DeviceFunction}}
\newcommand{\DProblem}{\dlfont{DeviceProblem}}
\newcommand{\DSH}{\dlfont{DomainSpecificHazard}}
\newcommand{\DSHShort}{\dlfont{DSpecificHazard}}
\newcommand{\Event}{\dlfont{Event}}
\newcommand{\Harm}{\dlfont{Harm}}
\newcommand{\Hazard}{\dlfont{Hazard}}
\newcommand{\HSituation}{\dlfont{HazardousSituation}}
\newcommand{\IManifest}{\dlfont{ImplementationManifest}}
\newcommand{\IManifestShort}{\dlfont{IManifest}}
\newcommand{\PProblem}{\dlfont{PatientProblem}}
\newcommand{\Probability}{\dlfont{Probability}}
\newcommand{\Risk}{\dlfont{Risk}}
\newcommand{\RLevel}{\dlfont{RiskLevel}}

\newcommand{\RSDA}{\dlfont{RiskSDA}}
\newcommand{\RSDAI}{\dlfont{RiskSDAI}}
\newcommand{\SAssurancce}{\dlfont{SafetyAssurance}}
\newcommand{\SDA}{\dlfont{SafeDesignArgument}}
\newcommand{\SDAShort}{\dlfont{SDA}}
\newcommand{\SDAI}{\dlfont{SDAI}}
\newcommand{\Severity}{\dlfont{Severity}}
\newcommand{\CRL}{\dlfont{CriticalRiskLevel}}

\newcommand{\cHarm}{\dlfont{causesHarm}}
\newcommand{\hARisk}{\dlfont{hasAnalyzedRisk}}
\newcommand{\hDComponent}{\dlfont{hasDeviceComponent}}
\newcommand{\hDContext}{\dlfont{hasDeviceContext}}
\newcommand{\hDFunction}{\dlfont{hasDeviceFunction}}
\newcommand{\hDProblem}{\dlfont{hasDeviceProblem}}
\newcommand{\hDSH}{\dlfont{hasDomainSpecificHazard}}
\newcommand{\hDSHShort}{\dlfont{hasDSHazard}}
\newcommand{\hEvent}{\dlfont{hasEvent}}
\newcommand{\hHarm}{\dlfont{hasHarm}}
\newcommand{\hHazard}{\dlfont{hasHazard}}
\newcommand{\hHSituation}{\dlfont{hasHazardousSituation}}
\newcommand{\hIManifest}{\dlfont{hasImplementationManifest}}
\newcommand{\hIManifestShort}{\dlfont{hasIManifest}}
\newcommand{\hIRL}{\dlfont{hasInitialRiskLevel}}
\newcommand{\hPProblem}{\dlfont{hasPatientProblem}}
\newcommand{\hPHazard}{\dlfont{hasParentHazard}}
\newcommand{\hPSituation}{\dlfont{hasParentSituation}}
\newcommand{\hPEvent}{\dlfont{hasPrecedingEvent}}
\newcommand{\hRRL}{\dlfont{hasResidualRiskLevel}}
\newcommand{\hRL}{\dlfont{hasRiskLevel}}

\newcommand{\hSAssurancce}{\dlfont{hasSafetyAssurance}}
\newcommand{\hSAssurancceShort}{\dlfont{hasSAssurance}}
\newcommand{\hSSDA}{\dlfont{hasSubSDA}}
\newcommand{\iMBy}{\dlfont{isMitigatedBy}}
\newcommand{\iPODComponent}{\dlfont{isPartOfDeviceComponent}}

\newcommand{\hProbability}{\dlfont{hasProbability}}
\newcommand{\hPOne}{\dlfont{hasProbability1}}
\newcommand{\hPTwo}{\dlfont{hasProbability2}}
\newcommand{\hSeverity}{\dlfont{hasSeverity}}

\newcommand{\ex}[1]{\exists #1.\top}
\newcommand{\gcialign}[2]{\ensuremath{#1 &\sqsubseteq #2}}

\newcommand{\dom}[1]{\mathit{dom}(#1)}
\newcommand{\ran}[1]{\mathit{ran}(#1)}
\newcommand{\tra}[1]{\mathit{tra}(#1)}

\DeclareFloatingEnvironment[fileext=lol, listname={List of Listings}, name=Listing]{listingi}

\newcommand{\EL}{\ensuremath{\mathord{\mathcal{E}\!\mathcal{L}}}\xspace}
\newcommand{\ELpp}{\ensuremath{\EL^{++}}\xspace}

\newcommand{\Inds}{\mathsf{N_I}}
\newcommand{\Cons}{\mathsf{N_C}}
\newcommand{\Rols}{\mathsf{N_R}}

\newcommand{\dland}{\sqcap}
\newcommand{\dlsub}{\sqsubseteq}

\newcommand{\dlint}{\ensuremath{\mathcal{I}}}

\newcommand{\underint}[1]{\ensuremath{{#1}^\dlint}}

\newcommand{\dldom}{\ensuremath{\underint{\Delta}}}
\newcommand{\dlfunc}{\ensuremath{\underint{\cdot}}}
\newcommand{\Dlint}{\ensuremath{\tuple{\dldom,\dlfunc}}}

\newcommand{\CName}{\ensuremath{\dlfont{A}}}
\newcommand{\RName}{\ensuremath{\dlfont{R}}}
\newcommand{\RNamesub}[1]{\ensuremath{\dlfont{R_{#1}}}}
\newcommand{\IName}{\ensuremath{\dlfont{a}}}
\newcommand{\JName}{\ensuremath{\dlfont{b}}}

\newcommand{\TBox}{\ensuremath{\mathcal{T}}\xspace}

\newcommand{\ABox}{\ensuremath{\mathcal{A}}\xspace}

\newcommand{\rassert}[3]{\mbox{\ensuremath{#3(#1,#2)}}}
\newcommand{\cassert}[2]{\mbox{\ensuremath{#2(#1)}}}

\newcommand{\ind}[1]{\dlfont{#1}}
\newcommand{\indid}[2]{$\dlfont{#1}_{\dlfont{#2}}$}
\newcommand{\indidb}[2]{(\indid{#1}{#2})}
\newcommand{\indb}[1]{(\ind{#1})}

\newcommand{\probcount}{\pi}
\newcommand{\sevcount}{\sigma}

\newcommand{\dlprob}[1][i]{\ind{p_{#1}}}
\newcommand{\dlsev}[1][i]{\ind{s_{#1}}}

\newcommand{\psonto}[2]{\ensuremath{\mathcal{K}^{{p}\textsc{-}\mkern-1mu{s}}_{#1,#2}}}
\newcommand{\psTBox}{\TBox_{\probcount,\sevcount}}
\newcommand{\psABox}{\ABox_{\probcount,\sevcount}}

\newcommand{\NPlus}{\mathbb{N}^{+}}
\newcommand{\Graph}{\ABox}
\newcommand{\VGraph}{\ensuremath{\Inds(\Graph)}}

\newcommand{\hasatleast}[1][n]{\mathord{\geq_{#1}}}
\newcommand{\hasatmost}[1][n]{\mathord{\leq_{#1}}}
\newcommand{\hasexactly}[1][1]{\mathord{=_{#1}}}
\newcommand{\hasexactlytop}[1]{\hasexactly[1]{#1}.\top}

\newcommand{\geval}[1]{\llbracket{#1}\rrbracket^{\Graph}}

\newcommand{\Constraints}{\mathcal{C}}
\newcommand{\Targets}{\mathcal{B}}
\newcommand{\cramalign}{\addtolength{\jot}{-1ex}}

\newcommand{\mysection}[1]{\section{#1}}
\newcommand{\mysubsection}[1]{\subsection{#1}}

\newcommand{\myemail}[1]{{\normalfont\ttfamily #1}}

\newcommand{\myparagraph}[1]{\vskip1pt\noindent\emph{#1}}

\newif\iflong
\ifdefined\longversion\longtrue\else\longfalse\fi

\begin{document}

\pagestyle{headings}

\newcommand{\ourtitle}{Supporting Risk~Management for Medical~Devices via the \href{http://w3id.org/riskman}{\riskman} Ontology~and~Shapes\iflong~(Preprint)\fi}

\begin{frontmatter}

      %\title{\href{https://en.wikipedia.org/wiki/A_Farewell_to_Arms}{A Farewell to Harms}:\\\ourtitle}
      \title{\ourtitle}
      \runningtitle{Supporting Risk~Management for Medical~Devices via \href{http://w3id.org/riskman}{\riskman}}

      \author[A]{\fnms{Piotr} \snm{Gorczyca}\orcidlink{0000-0002-6613-6061}}
      \author[A,D]{\fnms{Dörthe} \snm{Arndt}\orcidlink{0000-0002-7401-8487}}
      \author[A]{\fnms{Martin} \snm{Diller}\orcidlink{0000-0001-6342-0756}}
      \author[B]{\fnms{Jochen} \snm{Hampe}\orcidlink{0000-0002-2421-6127}}
      \author[C]{\fnms{Georg} \snm{Heidenreich}\orcidlink{0000-0001-7248-0566}}
      \author[A]{\fnms{Pascal} \snm{Kettmann}\orcidlink{0009-0009-9461-7952}}
      \author[A,D,E]{\fnms{Markus} \snm{Krötzsch}\orcidlink{0000-0002-9172-2601}}
      \author[A]{\fnms{Stephan} \snm{Mennicke}\orcidlink{0000-0002-3293-2940}}
      \author[A,D]{\fnms{Sebastian} \snm{Rudolph}\orcidlink{0000-0002-1609-2080}}
      \author[A,D]{\fnms{Hannes} \snm{Straß}\orcidlink{0000-0001-6180-6452}}
      \runningauthor{Gorczyca et al.}
      %\address[A]{Computational Logic Group, Institute of Artificial Intelligence}
      %\address[B]{Logic Programming and Argumentation Group, Institute of Artificial Intelligence}
      %\address[C]{Knowledge-Based Systems Group, Institute for Theoretical Computer Science}
      \address[A]{Faculty of Computer Science, TU Dresden, Germany; \myemail{firstname.lastname@tu-dresden.de}}
      \address[B]{Department of Medicine 1, University Hospital Dresden, TU Dresden, Germany; \myemail{jochen.hampe@uniklinikum-dresden.de}}
      \address[C]{Siemens Healthineers, Germany; \myemail{georg.heidenreich@siemens-healthineers.com}}
      \address[D]{ScaDS.AI Center for Scalable Data Analytics and Artificial Intelligence, Dresden/Leipzig, Germany}
      \address[E]{Centre for Tactile Internet with Human-in-the-Loop (CeTI), Dresden, Germany}
      \maketitle              % typeset the header of the contribution
      \begin{abstract}
            We propose the \riskman{} ontology and shapes for representing and analysing information about risk management for medical devices.
Risk management is concerned with taking necessary precautions to ensure that a medical device does not cause harms for users or the environment.
To date, risk management documentation is submitted to notified bodies (for certification) in the form of semi-structured natural language text.
We propose to use terms from the \riskman{} ontology to provide a formal, logical underpinning for risk management documentation, and to use the included SHACL constraints to check whether the provided data is in accordance with the requirements of the two relevant norms, i.e.\ ISO~14971 and VDE~Spec~90025.
\riskman is available at \riskmanurl.
%Our proposed methodology has the potential to save many person-hours for both manufacturers (when creating risk management documentation) as well as notified bodies (when assessing submitted applications for certification), and thus offers considerable benefits for healthcare and, by extension, society as a whole.
            % \metadata{Resource type}{OWL EL Ontology, SHACL Shapes}\\\metadata{License}{CC BY 4.0}
            % \quad~\metadata{URL}{\url{https://w3id.org/riskman/}}
            % \keywords{Risk management \and Safety assurance \and OWL EL \and SHACL}
      \end{abstract}


\end{frontmatter}

%%%%%%%%%%%%%%%%%%%%%%%%%%%%%%%%%%%%%%%%%%%%%%%%%%%%%%%%%%%%%%%%%%%%%%%%%%%%%%%%
\mysection{Introduction}

Medical devices typically are safety-critical, meaning their failure under certain conditions can lead to harm to humans or the environment.
To ensure that potential harms are minimized, legislation requires manufacturers of medical devices to provide a comprehensive justification that their product is acceptably safe.
In Europe, such legal requirements mainly stem from the European Union's Medical Device Regulation (EU MDR)~\cite{EUMDR}. %\footnote{Regulation (EU) 2017/745 of the European Parliament and of the Council of 5 April 2017 on medical devices, available at \url{http://data.europa.eu/eli/reg/2017/745/oj}.}
After its introduction in 2017, the original plan was that manufacturers have time until May 2024 to re-certify their devices under the new regulations.
This caused an immense backlog for notified bodies (organisations certifying medical devices in Europe, e.g., TÜVs in Germany), so the EU decided to extend the transition period until at least 2027 to avert a potential shortage of medical devices (and ramifications thereof)~\cite{mdrNewsURL}.%\footnote{\mbox{\url{https://www.medtechdive.com/news/EU-European-Parliament-MDR-extension/643064/}}}
The fundamental challenges of medical device certification are
(1) the sheer amount of information that notified bodies have to process (even for a single device), and
(2) the way this information is submitted to notified bodies, namely in the form of text.
In principle, device manufacturers have to argue that they have proactively, systematically, and thoroughly analyzed and mitigated the risks associated with their device as much as possible.
One way to do this is using a so-called \emph{assurance case} (AC), a structured argument supported by a body of evidence that jointly provide a compelling, comprehensible, and valid case that a system is acceptably safe for a given application in a given context.
% see Ryan Conmy et al. 2023, Section 2.1
Using assurance cases has a long(er) tradition in other areas of safety-critical systems, e.g.\ in aviation or nuclear power plants~\cite{SujanHKPJ16}, but the need for assurance cases for medical devices has been recognized long ago~\cite{WeinstockG09}.
The Food and Drug Administration of the U.S.~(FDA) has even provided a guidance document for one class of products (infusion pumps) due to an unusually large number of previous incidents~\cite{FDA14}.

As much as the introduction of assurance cases has helped advance safety management practices in the medical device industry, there is still a lot to be wanted.
Sujan et al.\xspace even claim that “safety management practices in healthcare [are] at present [\ldots] less mature than those in other safety-critical industries”~\cite[p.\,185]{SujanHKPJ16}.
% Sujan et al.~\cite{SujanHKPJ16} even claim that “safety management practices in healthcare [are] at present [\ldots] less mature than those in other safety-critical industries” (p185).
% Low maturity of safety management practices: it could be argued that safety management practices in healthcare at present are less mature than those in other safety-critical industries. The introduction of goal-based approaches to regulation including the development of safety cases may have the potential to contribute to a more structured and rigorous approach to safety management in this industry
We briefly note some identified shortcomings: %we address in this work.

(1) Since the intended recipient of a safety case is human (the auditor), a lot of work in safety engineering has focused on the \emph{presentation} of the argument, e.g.\ to avoid confirmation bias~\cite{ChowdhuryWPL20}.
% \item First, since the intended recipient of a safety case is a human (auditor), a lot of work in safety engineering has focused on the \emph{presentation} of the argument, e.g.\ to avoid confirmation bias~\cite{ChowdhuryWPL20}.
Several approaches to visualize assurance cases exist, e.g.\ the goal structuring notation~\cite{KellyM97}.
However, work on the \emph{representation} of safety arguments is rare, since applicable legislation typically requires a submission in the form of text.

(2) Due to the sheer volume of text submitted to notified bodies for certification, an auditor can never look at all relevant points in detail;
% \item Second, due to the sheer volume of text submitted to notified bodies for certification, an auditor can never look at all relevant points in detail;
furthermore, auditors typically spend a considerable proportion of their time navigating through documents (in the best case using text search in a document viewer).
%Both involved parties are in dire need of better software support: auditors would rather spend their time evaluating the actual risk management, and manufacturers clearly would benefit immensely from reduced duration of certification.
Tedious, manual examination of submissions is necessary to identify missing or incorrect information.
Significant time savings are offered when the initial inspection can be (semi-)automated and when improved navigation through e.g.\ semantics-enhanced document search is available.

(3) Reusing parts of safety arguments is a widely recognized problem~\cite{MartinBLW14,RuizJEVL17}.
% \item Third, re-using parts of safety arguments is a widely recognised problem~\cite{MartinBLW14,RuizJEVL17}.
The simplest instance of re-use happens when a manufacturer continues development of a certified device and wants to certify the “updated version”:
application lifecycle management tools help identify the delta between two device versions and risk managers can restrict attention on how that difference affects risks and mitigations.
% Harder instances of re-use are equally pervasive and important, e.g.\ when a manufacturer wants to use a part of a device in a novel device of a different type, or when a manufacturer buys parts (like integrated circuits) from component suppliers and needs to use assurances about the parts in order to claim overall device safety.
Harder instances of re-use are equally pervasive and important.
For example, this occurs when a manufacturer wants to use some device part in a novel device of a different type.
It also arises when a manufacturer buys parts (like integrated circuits) from component suppliers and uses assurances about those parts to claim overall device safety.
% maybe also cite 35, 36 from the Ruiz17 paper
We even propose to go beyond this and technically enable re-use \emph{across manufacturers}, which is especially relevant for small/medium enterprises that may lack staff or experience in risk management.

%\noindent Thus there is substantial demand for intelligent software support in risk management, both on the side of manufacturers as well as from notified bodies.\refreq{r}{3}
Therefore, any approach toward a more structured representation of safety cases, allowing for automated checks and facilitating re-use, would appear to be highly welcome by all involved parties.
% one-sentence paragraphs considered bad style
Consequently, in this paper, we propose to use logical modelling (more precisely the \emph{web ontology language}, OWL) %(in description logic terms, an ABox)
to represent risk management documentation, and the Shapes Constraint Language (SHACL) to check those representations for conformance with a set of requirements, e.g., whether all identified risks have an associated mitigation.
% To this end, we introduce the \riskman ontology and shapes.
While such straightforward checks could clearly also be achieved by custom software, the presence of a logical inference step before the constraint checking step is a significant advantage of our approach, as we shall demonstrate later in the paper.
Thus as our main contribution we introduce the \riskman ontology and shapes.

%Taken together, the \riskman ontology and its shapes constraints %, both available at \riskmanurl, constitute the resource we contribute in this paper, which is intended to be \riskman's canonical citation.
As far as we are aware, using OWL and SHACL is a novel approach to representing and reasoning about risk management documentation.\refreq{i}{1}
The requirements to be checked are simple and syntactic at the moment, but in view of communication with domain experts from notified bodies and manufacturers, we still expect our approach to be a major step forward as manufacturers (and notified bodies) can expect to the first round of conformity assessment being fully automated.\refreq{i}{2}
We would like to emphasize that our approach does not aim at assessing the adequacy and correctness of the implemented mitigations themselves (such as whether an insulation thickness of 0.5\,mm is adequate to lower the risk of electric shock). It stands to reason that such an analysis is presumably AI-complete and would at least require vast amounts of background knowledge.\refreq{r}{6}
Apart from facilitating tasks in this important application domain, our ontology and shapes are also of independent interest to the Semantic Web community, as the combination of logical inference and SHACL constraint checking is a topic that has garnered significant research interest lately~\cite{AhmetajOOS23,ParetiKNS19,demeesterswj2021}.\refreq{i}{5}
Finally, the fact that our approach relies on standardized and well-supported formalisms unlocks the adoption of Semantic Web technologies in general.\refreq{i}{7}
Employing these technologies enables \riskman users to benefit from existing infrastructure of tooling:
Risk reports can easily be queried or enhanced with additional information about the use context of a concrete device  or  data about particular patient problems by simply using other ontologies and data~\cite{cc,el2018snomed,ncitURL}.

The rest of the paper is organized as follows:
In the next section, we give an overview on related work. %, mostly ontologies involving notions of risk, but also software-based solutions for safety case management.
Afterwards (\Cref{sec:background}), we introduce the state of the art of risk management and the basic notions introduced by the norms upon which we build our ontology, both in general terms and with an illustrative running example.
\Cref{sec:ontology} then introduces the \riskman ontology and its associated shape constraints, explaining how we intend them to be used in risk management, and showcasing them with the running example.
In \Cref{sec:discussion} we conclude with a discussion of potential future work.

%%%%%%%%%%%%%%%%%%%%%%%%%%%%%%%%%%%%%%%%%%%%%%%%%%%%%%%%%%%%%%%%%%%%%%%%%%%%%%%%
\mysection{Related Work}
\label{sec:related-work}

We reviewed the literature for
(1) ontologies on (a) medical devices or (b) notions of \emph{risk}, with a special focus on an intended use for automated reasoning (in particular using SHACL constraints),
and (2) academic reports on software tools for risk management.

\subsection{Conceptual Work and Ontologies}

Fenz et al.~\cite{FenzPH16,FenzN18} created an ontology for the information security standard ISO~27002.
While they do not deal with risk management (or medical devices), they also use their ontology along with a reasoner to infer information about compliance with the standard's requirements.
Reasoning results are then interpreted by humans or by a tailor-made software tool.
% in our approach, the SHACL constraints take over this role to additionally streamline the overall process.
%
Uciteli et al.~\cite{UciteliNTSSFKSS17} provided the \emph{Risk Identification Ontology}, which defines notions of \emph{risk} and \emph{adverse situation} and embeds them into the top-level ontology GFO (General Formal Ontology)~\cite{Herre10,HerreHBHLM06}.
However, they work in a more process-oriented setting with special focus on risk identification in %perioperative environments, that is, time periods before, during, and after surgical procedures.
time periods surrounding surgical procedures.
In that setting, risks cannot be mitigated beforehand, and so \emph{risk management} -- our main focus -- is not within their scope.
Kim et al.~\cite{KimPLL19} presented a process integration ontology for medical software developers with a focus on medical devices, combining notions from \mbox{IEC~60601-1}, IEC~62304, and, notably, ISO~14971.
The resulting integrated ontology is however mainly designed to help developers comply with the involved standards;
the ontology was not developed with logical reasoning as explicit intended use case.
%
% BEGIN OPTIONAL
Aziz et al.~\cite{AzizAK19} developed a \emph{Hazard Identification Ontology}, involving notions of \emph{Hazard} and \emph{Events}.
Their focus was however on identifying risks rather than mitigating them, especially in scenarios of fire, explosions, or toxicity. %, where in this work we deal with risk management for medical devices.
% END OPTIONAL
%
Schütz et al.~\cite{SchutzFW20} created an ontology for medical devices in Germany, albeit more broadly targeting devices' manufacturers, operators, and legal procedures from an outside perspective with the aim of general semantic interoperability, and based on a legal framework that has since been superseded by the EU~MDR.
Single et al.~\cite{SingleSD20} presented an ontology for Hazard and Operability (HAZOP), a methodology for scenario-based hazard evaluation, therein defining notions of \emph{deviation}, \emph{cause}, \emph{effect}, \emph{consequence}, and \emph{safeguard}.
Their aim was however to create HAZOP worksheets (to be used by human operators) automatically.
McDonald et al.~\cite{McDonald21} mention how they use the “ARK Mindful Governance of operational risk formal ontology” to annotate textual risk-analysis data to make it amenable to machine processing (p12).
They however give no details on the ontology or its possible inferences.
Alanen et al.~\cite{AlanenLMPAHT22} provided a comprehensive risk assessment ontology including the notions \emph{risk}, \emph{risk level}, and \emph{risk control}, which is harmonized between (and intended for use across) safety, security, and dependability.
%It extended earlier work~\cite{} on a risk assessment ontology for safety. % only technical report
The intended use of their ontology is to “support the creation of a structured work product storage with traceability links” in order to improve upon current practices with non-structured word-processing documents, albeit without using logical reasoning.
Golpayegani et al.~\cite{GolpayeganiPL22} propose the AI Risk Ontology (AIRO) for expressing information associated with high-risk AI systems based on the requirements of the EU's AI Act~\cite{AIAct} and the ISO~31000 series of standards.
AIRO contains classes for \emph{risk}, \emph{risk source}, \emph{event}, \emph{consequence}, and \emph{impact}, and Golpayegani et al.\ also present SHACL shapes for determining whether an AI system is high-risk.
This is perhaps the closest relative of \riskman, only for AI systems instead of medical devices.
In conclusion, while various ontologies including notions of \emph{risk} exist, they either do not implement \isonormcite or are not designed with logical reasoning in mind, and most of them do not use SHACL for conformance testing (in the sense of pre-certification).\refreq{i}{3}\refreq{i}{4}

\subsection{(Logic-Based) Software Tools for Risk Management}

Fujita et al.~\cite{FujitaMHSKI12} presented the D-Case Editor, a dependability-oriented assurance case editor supporting the goal structuring notation~\cite{KellyM97}, implemented as an Eclipse plugin and available at its GitHub repository~\cite{d-case_editorURL}.
However, its development has been discontinued. % (the last repository change was in 2015).
Rushby~\cite{Rushby05} and Cruanes et al.~\cite{CruanesHOS13} introduced the Evidential Tool Bus, a Datalog-based system for integrating the development of safety cases into system (and software) development.
In particular, their approach covers the management of claims (about the system to be developed) and how they are supported by evidence provided by other software tools.
While an implementation is available~\cite{ETBURL}, it is not maintained any longer. % (last change 2016).
Beyene and Ruess~\cite{BeyeneR18} later picked up that work and extended it by connecting to the Jenkins~\cite{JenkinsURL} %\footnote{Available at \url{https://jenkins.io}.}
continuous integration software, with an active repository~\cite{JenkinsURLGit}.
The focus of the Evidential Tool Bus is primarily on obtaining, maintaining, and managing pieces of evidence in a development setting.
As such, it can be an important and complementary addition to using the methodology we propose in this paper.
Since the issue of safety is of great concern to industry, there are also several proprietary software tools centring around safety assurance:
The UK company Adelard offers the \href{https://www.adelard.com/asce/}{Assurance and Safety Case Environment}.
Similarly, de~la~Vara et al.~\cite{delaVaraJMP19} reported on using the \href{https://www.reusecompany.com/v-v-studio-verification-and-validation}{V\&V~studio} by REUSE Software.
In conclusion, existing software solutions for risk management are proprietary or do not use formal logics, or they have other (but complementary) use cases.\refreq{i}{4}

%%%%%%%%%%%%%%%%%%%%%%%%%%%%%%%%%%%%%%%%%%%%%%%%%%%%%%%%%%%%%%%%%%%%%%%%%%%%%%%%
\mysection{Background}
\label{sec:background}

\begin{table}[t]
  \centering
  \caption{Most important terms and definitions from \isonorm and \vdespec.}
  \label{tab:terms_definitions}
  \scriptsize
  \begin{tabular}{p{3cm}|p{8.8cm}}
    % \hline
    \toprule
    \textbf{Term}                                                       & \textbf{Definition}                                                                                                                                                                      \\
    \hline
    Analyzed risk                                                       & \textit{Combination of one or more domain-specific hazard(s) with one hazardous situation and one harm with reference to a device context and a specification of an initial risk level.} \\
    \hline
    Controlled risk                                                     & \textit{Structured artifact that relates one analyzed risk with one or more SDA(s) and specifies a resulting residual risk.}                                                             \\
    \hline
    Domain specific hazard                                              & \textit{Structured artifact that centres around one hazard having the potential to cause one or more harm(s) in the context of a domain-specific function and component.}                \\
    \hline
    Harm (\isonorm)                                                     & \textit{Injury or damage to the health of people, or damage to property or the environment}                                                                                              \\
    \hline
    Hazard (\isonorm)                                                   & \textit{Potential source of harm.}                                                                                                                                                       \\
    \hline
    \raggedright Hazardous situation \ \ \ \ \ \ \ \ \ \ \ \ (\isonorm) & \textit{Circumstance in which people, property or the environment is/are exposed to one or more hazards.}                                                                                \\
    \hline
    Risk level                                                          & \textit{Combination of probability and severity.}                                                                                                                                        \\
    \hline
    Severity (\isonorm)                                                 & \textit{Measure of the possible consequences of a hazard.}                                                                                                                               \\
    \bottomrule
  \end{tabular}
  \vskip1pt
\end{table}

The \riskman ontology is based on the recent \vdespeccite, which proposes a structured format for digitalizing risk management files, as well as a machine-readable exchange format using HTML with RDFa~\cite{RDFa}, which annotates (some) HTML tags with Resource Description Framework (RDF) triples.
\vdespec, in turn, is based on \isonormcite, which ``specifies terminology, {\it principles}, and a {\it process} for {\it risk management} of {\it medical devices}''.
Throughout this section, we briefly recall the notions from \vdespec and \isonorm that are central to the \riskman{} ontology.
Some of the notions are provided in \Cref{tab:terms_definitions}.
A full list of definitions of the terms can be found in \vdespecciteo[Section~3].
The model of risk underlying both \isonorm and \vdespec (see Annex C of \isonorm) is that of a {\it hazard} leading, via a sequence of {\it events}, to a {\it hazardous situation}, which, in turn, results in a {\it harm}, as depicted below:
\vskip1ex

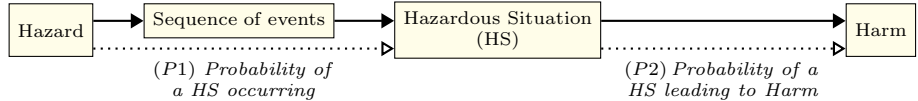
\noindent%
\begin{tikzpicture}[
    every node/.style={
            draw,
            align=center,
            font=\scriptsize,
            minimum height=0.7cm,
            fill=yellow!10,
            % rounded corners
        },
    nofill/.style={
            draw=none,
            fill=none
        },
    flow/.style={
    -{Triangle},%[open],
    %-{Stealth},auto,
    % color=green, 
    thick
    },
    prob/.style={
    -{Triangle}[open],
    dotted,
    thick
    }
    ]

    \node (hazard) {Hazard} ;
    \node[right=2.5cm of hazard.north, anchor=north, minimum height=0.25cm] (ev) {Sequence of events} ;
    \node[right=4cm of hazard] (hsit) {Hazardous Situation\\(HS)} ;
    \node[right=3.25cm of hsit] (hrm) {Harm} ;

    % flow
    \draw[->, flow] ([yshift=3.5pt] hazard.east) -- ([yshift=3.5pt] hazard.east -| ev.west) node[nofill, midway, right] {};
    \draw[->, flow] (ev.east) -- (ev.east -| hsit.west) node[nofill, midway, right] {};
    \draw[->, flow] ([yshift=3.5pt] hsit.east) -- ([yshift=3.5pt] hsit.east -| hrm.west) node[nofill, midway, right] {};

    % probabilities
    \draw[prob] ([yshift=-7pt] hazard.east) -- ([yshift=-7pt] hazard.east -| hsit.west) node[nofill, midway, below, yshift=-1pt] {(\Pone) \textit{Probability of}\\\textit{a HS occurring}};
    \draw[prob] ([yshift=-7pt] hsit.east) -- ([yshift=-7pt] hsit.east -| hrm.west) node[nofill, midway, below, yshift=-1pt] {\!(\Ptwo)\,\textit{Probability of a}\\\textit{HS leading to Harm}};

\end{tikzpicture}

\noindent According to this view, the risk analysis that manufacturers of medical devices must undertake involves compiling a list of known and foreseeable combinations of hazard, events, hazardous situation, and harm.
A further central aspect of risk assessment is estimating (initial) {\it risk levels} by associating a {\it probability of occurrence} and a {\it severity} to each harm.
The probability is often split into probabilities \Pone and \Ptwo, with the overall probability then being \mbox{$P=\Pone\cdot\Ptwo$}.
Building on \isonorm, \vdespec requires manufacturers to document risk analysis results as a list of {\it analyzed risks}, including related {\it harm}, {\it patient problem}, {\it device context}, the {\it hazardous situation} caused by preceding {\it event(s)}, the assessed {\it initial risk level}, and the associated {\it domain specific hazards}. These, in turn, further specify the underlying {\it hazard} through details on the {\it device problem}, {\it device function}, and {\it device component}.
The “problem” classes enable referencing terminology from the International Medical Device Regulators Forum (IMDRF)~\cite{IMDRF20AET}, specifically Annex E for patient problems and Annex A for device problems.
The notion of a {\it device} is absent in \vdespec and the \riskman ontology, since each risk management file addresses a single medical device. However, extending the ontology to include a {\it device} concept and link related entities would enable more comprehensive queries.
\vdespec does not provide requirements on the values used to specify probabilities or the severity either;
in particular, probabilities are not probabilities in the mathematical sense.
Rather, these values are interpreted as magnitudes within a scale that is used by the manufacturing company producing the risk management file (cf.~\Cref{sec:probabilities-severities}).

Risk assessment by manufacturers primarily aims at risk control, i.e.\ minimizing risks as much as possible~\cite{EUMDR}.
To document the mitigation strategies devised for the different risks, \vdespec proposes the use of {\it safe design arguments} (SDAs).
These, although building on the notion of assurance cases~\cite{WeinstockG09}, avoid some complexity in their use~\cite{Kelly03,Leveson11,Graydon15,Graydon17} by only requiring a considerably simplified structure.
Specifically, SDAs encode structured arguments, showing that a certain risk has been mitigated, as trees.
Thus, each SDA can have one or more sub-SDAs (its children), which serve to substantiate a claim made in the parent SDA.
Moreover, SDAs need ultimately be based on {\it safe design argument implementations} (SDAIs), i.e.\ all leaf SDAs need to be SDAIs.
These include not only a claim but also an {\it implementation manifest},
which gives detailed information on how the claim has been implemented and points to concrete evidence (e.g.\ additional documentation or a specific line of software source code) to support this.

\vdespec further distinguishes between {\it risk SDAs} and {\it assurance SDAs}, with assurance SDAs referring to some state-of-the-art \textit{safety assurance} -- e.g.\ %a section of 
a section of a norm mentioning a way of handling a risk.
All other SDAs are risk SDAs, while children of assurance SDAs must also be assurance SDAs.
%\vdespec requires SDAs in risk management files to be embedded within {\it controlled risks}, which, apart from including the analysed risks they ``control'', also refer to {\it residual risk levels}, i.e.\ probability and severity of the risk remaining after risk control.%
An SDA tree is embedded within a {\it controlled risk}, which, apart from referring to the analyzed risk it ``controls'', also indicates a {\it residual risk level}, i.e.\ probability and severity that remain after risk control.
\begin{example*}
  \label{ex:running}
  Insulin infusion pumps
  %a paradigmatic example for assurance cases in risk management~\cite{FDA14}
  aid in regulating blood glucose levels, especially of patients with diabetes, by administering fast-acting insulin via a catheter inserted beneath the skin. Based on the risk assessment for a generic infusion pump by Zhang et al.~\cite{zhangJJ10,zhangJJR11}, \Cref{fig:rmf} shows %data about 
  a controlled risk and associated SDA that can be extracted from a risk management file that follows \vdespec.
  Following~Zhang et al.~\cite[entry 4.3.9 in Table 4 in the appendix]{zhangJJ10}, the risk stems from an ``alarm malfunction'' hazard (indicated by \ind{hz} in~\Cref{fig:rmf}).
  It is further specified within a domain specific hazard (\ind{dsh}) with related information on the device component (\ind{dcm}), its function (\ind{df}) and problem (\ind{dp}).
  Specifically, the vibration mechanism of the non-audio alarm integrated into the pump may fail (event \indid{ev}{1}) under normal operating conditions (device context \ind{dcx}).
  Then, the patient may not become aware of an issue (event \indid{ev}{2}), which can lead to the patient receiving less insulin (hazardous situation \ind{hs}) and the patient losing consciousness (harm \ind{hr}).
  % Apart from the information for the domain specific hazard \indb{dsh} required by VDE Spec, the \riskman ontology also allows referring to terminology for medical device and patient problems put forward by the International Medical Device Regulators Forum (IMDRF)~\cite{IMDRF20AET} (fields labelled with \ind{dp} and \ind{pp}, respectively).
  The SDA (\indid{sd}{0}, based on the work of Zhang et al.~\cite[Table 3]{zhangJJR11}) consists of three sub-SDAs and expresses that there are alternative means of alerting the patient.
  The first sub-SDA \indidb{sd}{1} expresses that the alarm condition is also indicated through visual signals;
  the second sub-SDA \indidb{sd}{2} indicates that this notification is recurring;
  the third sub-SDA \indidb{sd}{3} expresses that there is also an additional audio alarm that will start unless the patient acknowledges the vibration or blinking.
  Moreover, according to sub-SDA \indidb{sd}{5} the audible signal is in accordance with regulations, here the assurance is IEC 60601 \indb{sa}.
  Thus, sub-SDA \indidb{sd}{5} is the only assurance SDA; %(indicated by the pink colour);
  all other SDAs are risk SDAs. % (purple).
  On the other hand, as required by VDE Spec, each leaf SDA is an SDAI, with their implementation manifests (\indid{im}{1},\indid{im}{2},\indid{im}{4},\indid{im}{5}) pointing to details and documentation.%
\end{example*}%

\begin{figure}
    \centering
    \resizebox{0.9\textwidth}{!}{
        \begin{tikzpicture}
            \tikzset {
                basic node/.style={
                        draw,
                        font=\scriptsize,
                        align=center,
                        rounded corners,
                        minimum width=1.1cm,
                        fill=yellow!20
                    }}

            % DSH    
            \node[basic node, minimum width=4cm] (dcomponent) at (0,0) {\textbf{Device component}:\\Non-audio alarm \indb{dcm}};
            \node[basic node, minimum width=4cm] (dfunction) at (0,-1) {\textbf{Device function}:\\Alarm \indb{df}};
            \node[basic node, minimum width=4cm] (dfunction) at (0,-2) {\textbf{Device problem}: Defective\\Alarm (IMDRF A160106) \indb{dp}};
            \node[basic node, minimum width=4cm] (hazard) at (0,-3) {\textbf{Hazard}: Non-audio \\alarm malfunction \indb{hz}};

            \node[basic node, minimum width=3cm] (ppro) at (4,0) {\textbf{Patient problem}:\\Loss of consciusness\\(IMDRF E0119) \indb{pp}};

            \node[basic node] (harm) at (6.5,0) {
                \textbf{Harm}: Loss\\
                of consciou-\\
                sness \indb{hr}};

            \node[basic node, minimum width=3cm] (dcontext) at (4,-1.2) {\textbf{Device context}:\\Normal use \indb{dcx}};
            \node[basic node, minimum width=3cm] (hazsit) at (4,-2.25) {\textbf{Hazardous situation}:\\Underdose \indb{hs}};
            \node[basic node, minimum width=3cm] (event) at (4,-3.63) {\textbf{Event}:\\Vibration mechanism\\fails \indidb{ev}{1} $\rightarrow$ Vibration\\cannot be felt \indidb{ev}{2}};

            % IRL
            \node[basic node] (sev) at (6.5,-1.2) {\textbf{Sev}:\\IV \indidb{s}{4}};
            \node[basic node] (prob1) at (6.5, -2.2) {\textbf{Prob 1}:\\V\indidb{p}{5}};
            \node[basic node] (prob2) at (6.5,-3.2) {\textbf{Prob 2}:\\IV\indidb{p}{4}};

            % IRL
            \node[basic node] (rsev) at (8.45,0) {\textbf{Sev}:\\IV\indidb{s}{4}};
            \node[basic node] (rprob) at (8.45,-1) {\textbf{Prob}:\\III\indidb{p}{3}};

            \node[basic node, fill=blue!10] (sda0) at (8.45,-4.2) {\textbf{SDA}:\\\indidb{sd}{0}};

            % DSH
            \node[draw,myborder,fit=(dcomponent)(dfunction)(hazard),inner sep=5pt,label={[font=\scriptsize]south:\textit{Domain Specific Hazard}~\indb{dsh}}] {};
            % IRL
            \node[draw,myborder,fit=(sev)(prob1)(prob2),yshift=-2pt,inner sep=5pt,label={[font=\scriptsize, text width=1.5cm, align=center]south:\textit{Initial Risk Level}~\indb{irl}}] {};
            % RRL
            \node[draw,myborder,fit=(rsev)(rprob),inner sep=5pt,label={[font=\scriptsize, text width=1.5cm, align=center]south:\textit{Residual Risk Level}~\indb{rrl}}] {};
            % ARisk
            \node[draw,myborder,fit=(dcomponent)(event)(harm),inner xsep=8pt, inner ysep=10pt, yshift=-5pt, xshift=-3pt, label={[font=\scriptsize]south:\textit{Analysed Risk}~\indb{ar}}](arisk){};
            % CRisk
            \node[draw,myborder,fit=(arisk)(sda0),inner xsep=7pt, inner ysep=8pt, yshift=-4pt, xshift=2pt, label={[font=\scriptsize]south:\textit{Controlled Risk}~\indb{cr}}] {};
        \end{tikzpicture}}%
    \\\resizebox{0.9\textwidth}{!}{
        \begin{forest}
            forked edges,
            for tree={align=center,base=bottom,draw,font=\scriptsize,edge={->},rounded corners,scale=0.9},
            im/.style={fill=green!10},
            sa/.style={fill=purple!15},
            sda/.style={fill=blue!10},
            asda/.style={fill=brown!15},
            [\textbf{SDA}:\\Alternative alerting when vibration mechanism of non-audio alarm fails~\indidb{sd}{0}, sda,
            [\textbf{SDA}:Additional\\visual (blinking)\\signal \indidb{sd}{1},sda,name=sda1
            [\textbf{IM}: Sec. 10.3 \\ of Alarm\\report \indidb{im}{1},im,name=im1]],
            [\textbf{SDA}: Notification\\ recurs every\\ $X$ minutes \indidb{sd}{2},sda,name=sda2
            [{\textbf{IM}: $X\eqdef 0.5$, \\ Sec. 10.7 \\ of Alarm\\report \indidb{im}{2}},im]],
            [\textbf{SDA}: Additio-\\nal audio\\alarm \indidb{sd}{3},sda,name=asda3
            [\textbf{SDA}: Audio\\signal if vibration\\signal not\\acknowledged \indidb{sd}{4},sda,name=asda4
            [\textbf{IM}: Sec. \\ 10.11  of Alarm\\report \indidb{im}{4},im]],
            [\textbf{SDA}:\\Audible signal is\\at least X db \\ at Y m  \indidb{sd}{5},asda,name=asda5
            [{\textbf{IM}:  $X\eqdef 45$, \\ $Y\eqdef 1$ ,  Sec. 5.3 \\ of  Loudspeaker \\ test \indidb{im}{5}},im,name=im5]],
            [\textbf{SA}: IEC\\60601 \indb{sa}, no edge, sa,name=sa5,yshift=6pt]
            ],
            ]
            \draw[->] (asda5) to (sa5);
            \node[draw,myborder,fit=(sda1)(im1),label={[font=\scriptsize]south:\textit{SDAI}}] {};
            \node[draw,myborder,fit=(asda5)(sa5),label={[font=\scriptsize, text width=1.8cm, xshift=25pt, yshift=2pt, align=right]south:\textit{Assurance SDA}}] {};
            \node[draw,myborder,fit=(asda5)(sa5)(im5),inner sep=5pt,label={[font=\scriptsize]south:\textit{Assurance SDAI}}] {};
        \end{forest}
    }
    \caption{Graphical representation of the data of a controlled risk (top) and associated SDA (bottom) provided within a risk management file for an infusion pump described in~\Cref{ex:running}. Dashed boxes illustrate how related elements are grouped -- e.g. the largest box shows that an Analyzed Risk, Residual Risk Level, and SDA together form a Controlled Risk.
        In the bottom part, colours distinguish elements of different classes:
        blue (Risk SDA);
        brown (Assurance SDA);
        green (Implementation Manifest);
        pink (Safety Assurance).
        An abbreviation in parentheses next to each element shows its unique identifier (used later in~\Cref{fig:abox:graph} for reference).
        Arrows arrange the elements from the bottom part into a tree.}
    \label{fig:rmf}
\end{figure}

%%%%%%%%%%%%%%%%%%%%%%%%%%%%%%%%%%%%%%%%%%%%%%%%%%%%%%%%%%%%%%%%%%%%%%%%%%%%%%%%
\mysection{The \texorpdfstring{\riskman}{Riskman} Ontology and Shapes: Overview and Usage}
\label{sec:ontology}

% Define the Dog ear shape

% taken from manual
\makeatletter
\pgfdeclareshape{document}{
    \inheritsavedanchors[from=rectangle] % this is nearly a rectangle
    \inheritanchorborder[from=rectangle]
    \inheritanchor[from=rectangle]{center}
    \inheritanchor[from=rectangle]{north}
    \inheritanchor[from=rectangle]{south}
    \inheritanchor[from=rectangle]{west}
    \inheritanchor[from=rectangle]{east}
    % ... and possibly more
    \backgroundpath{% this is new
        % store lower right in xa/ya and upper right in xb/yb
        \southwest \pgf@xa=\pgf@x \pgf@ya=\pgf@y
        \northeast \pgf@xb=\pgf@x \pgf@yb=\pgf@y
        % compute corner of ‘‘flipped page’’
        \pgf@xc=\pgf@xb \advance\pgf@xc by-10pt % this should be a parameter
        \pgf@yc=\pgf@yb \advance\pgf@yc by-10pt
        % construct main path
        \pgfpathmoveto{\pgfpoint{\pgf@xa}{\pgf@ya}}
        \pgfpathlineto{\pgfpoint{\pgf@xa}{\pgf@yb}}
        \pgfpathlineto{\pgfpoint{\pgf@xc}{\pgf@yb}}
        \pgfpathlineto{\pgfpoint{\pgf@xb}{\pgf@yc}}
        \pgfpathlineto{\pgfpoint{\pgf@xb}{\pgf@ya}}
        \pgfpathclose
        % add little corner
        \pgfpathmoveto{\pgfpoint{\pgf@xc}{\pgf@yb}}
        \pgfpathlineto{\pgfpoint{\pgf@xc}{\pgf@yc}}
        \pgfpathlineto{\pgfpoint{\pgf@xb}{\pgf@yc}}
        \pgfpathlineto{\pgfpoint{\pgf@xc}{\pgf@yc}}
    }
}
\makeatother

The \riskman ontology was developed taking into account requirements and conceptual design choices laid out by a consortium of domain experts in medical devices and risk management, ontology engineers, and software developers.
For ontology development, we used the Linked Open Terms (LOT) methodology~\cite{Poveda-Villalon2022}, an ontology development framework that builds upon the NeOn methodology~\cite{SurezFigueroa2015TheNM}.
%This methodology structured the development into the following activities:
It comprised the following:

\myparagraph{Requirement specification.} % HS: Only activate footnote back again in final version.
%\footnote{The list of contributors to that activity, in particular for the creation of \vdespeccite, goes beyond the list of authors of this paper. We report on that phase for the sake of completeness, but this article does not claim credit for that work.}
Studying relevant norms/standards and inspecting real medical devices' risk management files to gain a shared understanding of the required terms.
This was accompanied by several meetings within the consortium and workshops with domain experts from the field: risk managers, manufacturers, consultants, and notified bodies.
As a result, we compiled objectives and a glossary of terms, forming the basis for the conceptual ontology model.
In parallel, the \vdespeccite working group developed a technical submission format with constraints based on the conceptual model.

\myparagraph{Implementation.}
% Following the requirements, 
We encoded the ontology and shapes using OWL and SHACL.
% which we will discuss in more detail later in this section.
During development, we used the HermiT reasoner~\cite{shearer2008hermit} and the \emph{OntOlogy Pitfalls Scanner} (OOPS!)~\cite{poveda2014oops} to ensure the ontology's validity and consistency.

\myparagraph{Publication and Maintenance.}
The \riskman ontology, shapes, and documentation are published online with a permanent URL provided by w3id.
To assess the conformity of the ontology to the FAIR (Findable, Accessible, Interoperable, and Reusable) principles~\cite{wilkonson2016fair}, we used the \emph{Ontology Pitfall Scanner for the FAIR Principles} (FOOPS!)~\cite{foops2021}.
\riskman is maintained in a GitHub repository~\cite{RiskmanRepo} for issue tracking/version control.

% As previously noted, the approach expects risk documentation to be submitted as HTML files containing risk management data encoded in RDF triples via RDFa.
As previously noted, the approach expects risk documentation to be submitted as HTML files containing risk management data encoded as RDF triples.
An \emph{RDF distiller} extracts the RDF graph from the HTML by removing the markup and isolating the RDF triples, which are then used as input to an \EL~\emph{reasoner} alongside \riskman (and potentially additional ontologies).
% It is then extracted using an \emph{RDF distiller} and used as input to an \EL~\emph{reasoner} alongside with \riskman and potentially additional ontologies.
As a result,
% The output of this procedure, 
a materialized DL knowledge base is obtained, which is then validated against the \riskman SHACL constraints (and potentially additional constraints) utilizing a \emph{SHACL validator}.
The outcome is then communicated with a human-readable \emph{validation report}.\refreq{r}{5}
We provide a prototypical implementation of the validation pipeline together with examples showcasing validation results~\cite{RiskmanPipeline}.
The symbolical depiction below presents the architecture and data flow of the approach:  \\

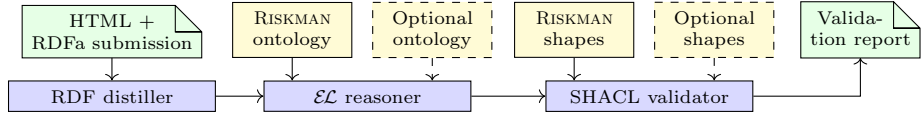
\noindent
\hspace*{-0.4ex}\begin{tikzpicture}[
        every node/.style={
                draw,
                align=center,
                font=\scriptsize,
                % rounded corners
            },
        pipeline/.style={
                minimum width=2.75cm,
                fill=blue!15
            },
        opt/.style={
                dashed
            },
        ontoshape/.style={
                text width=1.35cm,
                fill=yellow!20
            },
        io/.style={
                fill=green!10
            }
    ]

    \node[pipeline] (distiller) {RDF distiller};
    \node[pipeline] (reasoner) [right=18pt of distiller]  {\EL reasoner};
    \node[pipeline] (validator) [right=28pt of reasoner]  {SHACL validator};

    \node[shape=document, io] (input) [above=0.3cm of distiller]  {HTML +\\RDFa submission};

    \node[ontoshape] (ronto) [above=0.3cm of reasoner, xshift=-1cm]  {\riskman\\ontology};
    \node[opt, ontoshape] (aonto) [right=7.5pt of ronto]  {Optional\\ontology};

    \node[ontoshape] (rshape) [above=0.3cm  of validator, xshift=-1cm]  {\riskman\\shapes};
    \node[ontoshape, opt] (ashape) [right=7.5pt of rshape]  {Optional\\shapes};

    \node[shape=document, io] (output) [right=10 pt of ashape]  {Valida-\quad\qquad\\tion report};

    \draw[->] (input) -- (distiller);
    \draw[->] (ronto) -- (ronto.south|-reasoner.north) ;
    \draw[->, dashed] (aonto) -- (aonto.south|-reasoner.north)  ;
    \draw[->] (rshape) --  (rshape.south|-validator.north);
    \draw[->, dashed] (ashape) -- (ashape.south|-validator.north);

    \draw[->] (distiller) -- (reasoner) ;
    \draw[->] (reasoner) -- (validator) ;
    \draw[->] (validator.east) -| (output.south) ;
\end{tikzpicture}
\\

\noindent In the following subsections, we explain the approach in more detail and provide preliminaries necessary for understanding.
Afterwards, we present the actual \riskman{} ontology and shapes.
The section concludes with illustrating the approach on our running example and showing how it can be extended to further needs of interested parties.

%%%%%%%%%%%%%%%%%%%%%%%%%%%%%%%%%%%%%%%%%%%%%%%%%%%%%%%%%%%%%%%%%%%%%%%%%%%%%%%%

\mysubsection{The Description Logic \texorpdfstring{\EL}{EL}}
\label{sec:dl-el}
\noindent
Firstly, it is important to note that all the \riskman{} ontology can be represented within the description logic \ELpp~\cite{BaaderBL05,BaaderLB08}.
We therefore briefly recall its syntax, semantics, and aspects of reasoning that are relevant to our approach.

\ELpp's concept constructors and their semantics are recalled in the upper part of \Cref{tab:elpp:semantics};
the middle (lower) part shows the constructs allowed in a TBox (ABox).
\begin{table}[h]
    \centering
    \caption{Syntax and semantics of concept, TBox, and ABox expressions of \ELpp.}
    \label{tab:elpp:semantics}
    \setlength{\tabcolsep}{3pt} % Spacing between columns
    \noindent
    %\resizebox{0.95\linewidth}{!}{
    % \begin{tabular}{|l | c |  c|}
    \bgroup
    \renewcommand{\arraystretch}{0.9}
    \begin{tabular}{l  c  c}
        % \hline
        \toprule
        Name                                    & Syntax                                                   & Semantics                                                                                                                                      \\\midrule
        % Name                    & Syntax                                                   & Semantics                                                                                                                                       \\\hline
        individual name                         & $\IName\in\Inds$                                         & $\underint{\IName}\in\dldom$                                                                                                                   \\
        concept name                            & $\CName\in\Cons$                                         & $\underint{\CName}\subseteq\dldom$                                                                                                             \\
        role name                               & $\RName\in\Rols$                                         & $\underint{\RName}\subseteq\dldom\times\dldom$                                                                                                 \\
        top                                     & $\top$                                                   & $\dldom$                                                                                                                                       \\
        bottom                                  & $\bot $                                                  & $\emptyset$                                                                                                                                    \\
        nominal                                 & $\set{\IName}$                                           & $\set{\underint{\IName}}$                                                                                                                      \\
        conjunction                             & $C\dland D$                                              & $\underint{C}\cap\underint{D}$                                                                                                                 \\
        existential restriction                 & $\exists\RName.C$                                        & \hspace*{-1em}$\set{x\in\dldom\guard \exists y\in\dldom\colon\tuple{x,y}\in\underint{\RName} \mathbin{\,\&\,} y\in\underint{C}}$\hspace*{-1ex} \\[1pt]\midrule
        % existential restriction & $\exists\RName.C$                                        & $\underint{(\exists\RName.C)}\eqdef\set{x\in\dldom\guard \exists y\in\dldom\colon\tuple{x,y}\in\underint{\RName}\text{ and } y\in\underint{C}}$ \\[2pt]\hline
        \rule{-3pt}{8pt}
        range restriction                       & $\ran{\RName}\dlsub\CName$                               & $\underint{\RName}\subseteq\dldom\times\underint{\CName}$                                                                                      \\
        general concept inclusion\hspace*{-1ex} & $C\dlsub D $                                             & $\underint{C}\subseteq\underint{D}$                                                                                                            \\
        role inclusion axiom                    & $\RNamesub{1}\circ\dotsb\circ\RNamesub{k} \dlsub \RName$ & $\underint{\RNamesub{1}}\circ\dotsb\circ\underint{\RNamesub{k}}\subseteq\underint{\RName}$                                                     \\\midrule
        \rule{-3pt}{8pt}
        concept assertion                       & $\cassert{\IName}{\CName}$                               & $\underint{\IName}\in\underint{\CName}$                                                                                                        \\
        role assertion                          & $\rassert{\IName}{\JName}{\RName}$                       & $\tuple{\underint{\IName},\underint{\JName}}\in\underint{\RName}$                                                                              \\\bottomrule
        % RIA                     & $\RNamesub{1}\circ\dotsb\circ\RNamesub{k} \dlsub \RName$ & $\underint{\RNamesub{1}}\circ\dotsb\circ\underint{\RNamesub{k}}\subseteq\underint{\RName}$                                                      \\\hline
    \end{tabular}%}
    \egroup
    \vskip1pt
\end{table}
As usual for description logics, the semantics of \ELpp is defined via \define{interpretations} $\dlint=\Dlint$ with a non-empty \define{domain} $\dldom$ and an \define{interpretation function} $\dlfunc$.
%An interpretation $\dlint$ \define{satisfies} an assertion $\cassert{\IName}{\CName}$ ($\rassert{\IName}{\JName}{\RName}$) iff  ().
An interpretation $\dlint$ is a \define{model} of an ABox $\ABox$ (TBox $\TBox$) if it satisfies all elements of $\ABox$ ($\TBox$) as per \Cref{tab:elpp:semantics}.
An assertion $\alpha$ is \define{entailed} by $\TBox\cup\ABox$, written $\TBox\cup\ABox\models\alpha$, if every model of $\TBox\cup\ABox$ is a model of $\alpha$.
We also remark that RIAs (\ref{ria:hharm})--(\ref{ria:transitivity}) satisfy the syntactic restriction imposed by Baader et al.~\cite[Section~3]{BaaderLB08}.%, whence entailment of assertions can be decided in polynomial time.

There are two kinds of \EL reasoners, predominantly \emph{ontology classifiers}~\cite{KKS:ELK2013}.
The first kind implements %(logical)
\emph{tableau calculi} while the second is based on \emph{materialisation}.
% The latter kind of implementations fits well with our goals (i.e., assertion entailment and constraint checking).
Tools like ELK~\cite{KKS:ELK2013}, also distributed with ontology modelling frontends like Prot\'{e}g\'{e}~\cite{Musen15}, follow the materialisation-based approach by step-wise computing the relevant logical consequences of a given ABox
% (i.e., the risk management information) 
and an \ELpp ontology. %\footnote{More precisely, the “relevant” consequences in our case are all atomic concept assertions and role assertions involving concept, role, and individual names ocurring in the input ABox.}
% (i.e., \riskman{}). 
% Materialisation
In the context of \riskman{}, the “relevant” consequences are all concept and role assertions involving concept, role, and individual names occurring in the input ABox, where concept assertions must be atomic, that is, of the form $\CName(\IName)$ with $\CName\in\Cons$. %an \EL reasoner is used to complete the ABox (providing the risk management information), i.e.\ to establish which \riskman classes or object properties the individuals named in the ABox belong to.
The fact that no new individuals are required for this process is guaranteed by \riskman's design, which employs existential quantification only on the left-hand side of general concept inclusion axioms.\footnote{The optional add-on of \Cref{sec:probabilities-severities} has existential restriction on right-hand sides, but only involving nominal concepts, thus no new individuals need to be created during inference.}
The completion (addition of all relevant consequences) of a given ABox w.r.t.\ the \riskman ontology can therefore (also due to the satisfaction of the syntactic restriction by Baader et al.~\cite[Section~3]{BaaderLB08}) be computed in polynomial time~\cite{BaaderBL05,BaaderLB08},
% guarantees that the materialisation of any ABox against the set of \riskman axioms is decidable (actually even tractable), the procedure will always terminate and the set of all relevant logical consequences will be finite.
%The polynomial-time complexity of \ELpp reasoning~\cite{BaaderBL05,BaaderLB08},
which enables general purpose Datalog reasoners (e.g.\ Nemo~\cite{nemo24}) to implement \riskman ABox completion~\cite{CDK20:DatalogEL}.
Thus overall, a variety of optimized tools are at the disposal of potential \riskman users.
% In order to answer several queries (i.e., assertion entailments), tools like ELK~\cite{KKS:ELK2013}

\mysubsection{Classes, Properties, and Ontology Design Patterns}
\noindent
The main axioms of the \riskman~ontology are given in \Cref{fig:axioms}, including detailed superclass declarations, and relationships between properties via role inclusion axioms. %, role chains, or transitivity statements.
\riskman's classes and their general interrelationships are depicted in the schema diagram in \Cref{fig:schema}, where also domains and ranges of properties can be read off the edges.\refreq{q}{4}
\begin{figure}
    \smaller
    \cramalign
    \begin{align}
        \begin{split}\label{gci:arisk}
            \ex{\hDContext}\sqcap\ex{\hDSH} \sqcap \ex{\hHarm} &                   \\
            \sqcap~\ex{\hHSituation} \sqcap \ex{\hIRL}         & \sqsubseteq\ARisk
        \end{split}                             \\
        \gcialign{\SDA\sqcap\ex{\hSAssurancce}}{\ASDA} \label{gci:asda}                                    \\
        \gcialign{\SDAI\sqcap\ASDA}{\ASDAI} \label{gci:asdai}                                              \\
        \ex{\hARisk}\sqcap\ex{\hRRL} \sqcap \ex{\iMBy}               & \sqsubseteq\CRisk \label{gci:crisk} \\
        \ex{\hDComponent}\sqcap\ex{\hDFunction} \sqcap \ex{\hHazard} & \sqsubseteq\DSH \label{gci:dsh}     \\
        \gcialign{\ex{\hEvent}}{\HSituation} \label{gci:hsituation}                                        \\
        \gcialign{\ex{\hHarm}\sqcap\ex{\hRL}}{\Risk} \label{gci:risk}                                      \\
        \gcialign{\ex{\hProbability}\sqcap\ex{\hSeverity}}{\RLevel} \label{gci:rlevel}                     \\
        \gcialign{\RSDA}{\SDA} \label{gci:sda}                                                             \\
        \gcialign{\RSDA\sqcap\SDAI}{\RSDAI} \label{gci:rsdai}                                              \\
        \gcialign{\SDA\sqcap\ex{\hIManifest}}{\SDAI} \label{gci:sdai}                                      \\
        \gcialign{\hARisk\circ\hHarm}{\hHarm} \label{ria:hharm}                                            \\
        \gcialign{\hIRL}{\hRL} \label{ria:hrl:one}                                                         \\
        \gcialign{\hRRL}{\hRL} \label{ria:hrl:two}                                                         \\
        \begin{split}\label{ria:transitivity}
            \tra{\hPHazard}      & \quad\tra{\hPSituation} \\
            \tra{\iPODComponent} & \quad\tra{\hPEvent}
        \end{split}
    \end{align}%
    \caption{Main axioms of the \riskman ontology, i.e.\ those formalizing the definitions of \vdespec in \ELpp, with
        (\ref{gci:arisk})--(\ref{gci:sdai}) general concept inclusions (GCIs) and (\ref{ria:hharm})--(\ref{ria:transitivity}) role inclusion axioms (RIAs).
        Further axioms (subclass relationships, domain/range declarations, and disjointness axioms) can be read off \Cref{fig:schema}.
        The properties \hPHazard, \hPSituation, and \iPODComponent\xspace model hierarchies, while \hPEvent\xspace models temporal order in event chains; thus they all are defined as being transitive.
    }
    \label{fig:axioms}
\end{figure}

\begin{figure}[ht]
      \centering
      \begin{tikzpicture}
            \tikzset{
            every node/.style={draw, font=\tiny, fill=yellow!20},
            dsh col/.style={minimum width=2.2cm},
            edge label/.style={draw=none, font=\tiny, align=left, fill=none},
            hsit col/.style={minimum width=1.8cm},
            subClassOf/.style={-{Triangle}[open]},
            ObjectProperty/.style={-{Stealth},auto, ObjPro},
            }

            % dsh
            \node[minimum width=2.6cm] (dsh) {\DSH} ;
            \node[right=40pt of dsh] (haz) {\Hazard} ;
            \draw[->, ObjectProperty] ([yshift=3pt]haz.east) to [out=50,in=330,looseness=8] node[edge label, midway, right, align=left] {\dlfont{has}\\\dlfont{ParentHazard}} ([yshift=-3pt]haz.east);

            \node[dsh col, below right=17pt and 10pt of haz.west, anchor=west] (df) {\DFunction} ;
            \node[dsh col, below=17pt of df.west, anchor=west] (dp) {\DProblem} ;
            \node[dsh col, below=17pt of dp.west, anchor=west] (dcm) {\DComponent} ;
            \draw[->, ObjectProperty] ([xshift=20pt]dcm.south) to [out=300,in=250,looseness=4] node[edge label, midway, below, align=left] {\dlfont{isPartOf}\\\dlfont{Device}\\\dlfont{Component}} ([xshift=10pt]dcm.south);

            % DSH edges
            % DSH -> Hazard
            \draw[->, ObjectProperty] (dsh) -- node[edge label, above] {\hHazard} (haz);
            % DSH -> DFunction
            \draw[->, ObjectProperty] (dsh.south) ++(1.15cm,0) |- node[edge label, pos=0.75, above] {\hDFunction} (df.west);
            % DSH -> DProblem
            \draw[->, ObjectProperty] (dsh.south) ++(1.15cm,0) |- node[edge label, pos=0.75, above] {\hDProblem} (dp.west);
            % DSH -> DComponent
            \draw[->, ObjectProperty] (dsh.south) ++(1.15cm,0) |-  node[edge label, pos=0.75, above, yshift=-2pt] {\dlfont{hasDevice}\\\dlfont{Component}} (dcm.west);

            % ARisk
            \node[below=of dsh, minimum width=1.8cm] (ar) {\ARisk} ;
            % ARisk -> DSH
            \draw[->, ObjectProperty] ([xshift=20pt] ar.north) -- ([xshift=20pt] ar.north |- dsh.south) node[edge label, midway, left, align=right] {\dlfont{has}\\\dlfont{Domain}\\\dlfont{Specific}\\\dlfont{Hazard}};

            % Harm
            \node[below=17pt of dcm.west, anchor=west] (hrm) {\Harm} ;
            % DSH -> Harm
            \draw[->, ObjectProperty] (dsh.south) ++(1.15cm,0) |-  node[edge label, pos=0.75, above] {\cHarm} ([yshift=4pt] hrm.west);

            % Risk      
            \node[below left=2pt and 50pt of hrm] (risk) {\Risk} ;
            % Risk -> Harm
            \draw[->, ObjectProperty] ([xshift=3pt] risk.north) |- ([yshift=-3pt] hrm.west) node[edge label, below, pos=0.7] {\hHarm};

            % Arisk -> Risk   
            \draw[->, subClassOf] ([xshift=25pt] ar.south) -- ([xshift=25pt] ar.south |- risk.north) node[edge label, midway, right] {};

            % CRisk
            \node[below=100pt of dsh.west, anchor=west, minimum width=4cm] (cr) {\CRisk} ;
            % CRisk -> ARisk
            \draw[->, ObjectProperty] ([xshift=-20pt] cr.north) -- ([xshift=-20pt] cr.north |- ar.south) node[edge label, left, pos=0.2, align=right] {\dlfont{hasAna}-\\\dlfont{lyzedRisk}};
            % CRisk -> Risk
            \draw[->, subClassOf] ([xshift=2pt] cr.north) -- ([xshift=2pt] cr.north |- risk.south) node[edge label, midway, right] {};

            % Event
            \node[left=75pt of dsh] (ev) {\Event} ;
            % Event -> Event
            \draw[->, ObjectProperty] ([yshift=3pt]ev.west) to [out=140,in=210,looseness=8] node[edge label, midway, left, align=right] {\dlfont{has}\\\dlfont{PrecedingEvent}} ([yshift=-3pt]ev.west);

            % HSituation
            \node[hsit col, below=23pt of ev.west, anchor=west] (hsit) {\HSituation} ;
            % Hsit -> HSit
            \draw[->, ObjectProperty] ([yshift=3pt]hsit.west) to [out=140,in=210,looseness=8] node[edge label, midway, left, align=right] {\dlfont{has}\\\dlfont{ParentSituation}} ([yshift=-3pt]hsit.west);

            % PatientProblem
            \node[right=5pt of ev] (ppro) {\PProblem} ;

            % Device Context
            \node[hsit col, below=17pt of hsit.east, anchor=east] (dcx) {\DContext} ;
            % Risk Level
            \node[hsit col, below=17pt of dcx.east, anchor=east] (rl) {\RLevel} ;
            % Probability
            \node[minimum width=1.5cm, below=17pt of rl.west, anchor=west] (prob) {\Probability} ;
            % Severity
            \node[minimum width=1.5cm, below=17pt of prob.west, anchor=west] (sev) {\Severity} ;

            % Hazardous situation -> Event
            \draw[->, ObjectProperty] ([xshift=-30pt] hsit.north) -- ([xshift=-30pt] hsit.north |- ev.south) node[edge label, midway, left] {\hEvent};

            % Risk Level
            % Risk Level -> Probability
            \draw[->, ObjectProperty] (rl.west) -|  node[edge label, pos=0.8,  left] {\hProbability,\\\hPOne,\\\hPTwo} ++(-10pt,-17pt) -- (prob.west);

            % Risk Level -> Severity
            \draw[->, ObjectProperty] (rl.west) -|  node[edge label, pos=0.95,  left] {\hSeverity} ++(-10pt,-34pt) -- (sev.west);

            % ARisk -> Device context
            \draw[->, ObjectProperty] ([yshift=1pt] ar.west) --  node[edge label, above, pos=0.5] {\dlfont{hasDeviceContext}} (dcx.east);
            % ARisk -> Hazardous Situation
            \draw[->, ObjectProperty] (ar.north) ++(-0.25cm,0) |-  node[edge label, above, pos=0.75, yshift=-1pt] {\hHSituation} (hsit.east);
            % ARisk -> Patient Problem
            \draw[->, ObjectProperty] (ar.north)
            ++(-0.25cm, 0) % first horizontal movement
            |- ++(0, 0.75cm) % first vertical movement (up)
            -| ++(-2cm, 0) % second horizontal movement (left)
            -| node[edge label, right, pos=0.065, yshift=3.5pt] {\hPProblem} (ppro); % second vertical movement to connect

            % ARisk -> Risk Level
            \draw[->, ObjectProperty] (ar.south) ++(-0.5cm,0) |-  node[edge label, above, pos=0.75, yshift=-1pt] {\hIRL} ([yshift=3pt] rl.east);
            % Risk -> Risk Level
            \draw[->, ObjectProperty] (risk.north)++(-0.25cm,0) |-  node[edge label, below, pos=0.75, yshift=1pt] {\hRL} ([yshift=-3pt] rl.east);
            % CRisk -> Risk Level
            \draw[->, ObjectProperty] (cr.west) -|  node[edge label, below, pos=0.75, right] {\dlfont{hasResidual}\\\dlfont{RiskLevel}} ([xshift=20pt] rl.south);

            % SDA part
            \node[minimum width=1.3cm, below right=35pt and -70pt of cr] (sda) {\dlfont{SDA}} ;
            \draw[->, ObjectProperty] ([xshift=0pt]sda.north) to [out=130,in=50,looseness=7] node[edge label, midway, above, align=left] {\parbox{4mm}{\dlfont{has}\\[-2pt]\dlfont{Sub}\\[-2pt]\dlfont{SDA}}} ([xshift=6pt]sda.north);

            \node[minimum width=1.3cm, below=10pt of sda] (sdai) {\SDAI} ;
            \node[minimum width=2cm, right=22pt of sda] (asda) {\ASDA} ;
            \node[minimum width=2cm, above=25pt of asda.east, anchor=east] (sa) {\SAssurancce} ;
            \node[minimum width=2cm, below=10pt of asda] (asdai) {\ASDAI} ;
            \node[minimum width=1.3cm, left=of sda] (rsda) {\RSDA} ;
            \node[above=25pt of rsda.west, anchor=west] (im) {\IManifest} ;
            \node[minimum width=1.3cm, below=10pt of rsda] (rsdai) {\RSDAI} ;

            % CRisk -> SDA
            \draw[->, ObjectProperty] ([xshift=20pt] cr.south) -- ([xshift=20pt] cr.south |- sda.north) node[edge label, midway, right, pos=0.5, xshift=-2pt] {\parbox{4mm}{\dlfont{is}\\[-1pt]\dlfont{Mitigated}\\[-1pt]\dlfont{By}}};
            % SDA -> Implementation Manifest
            \draw[->, ObjectProperty] ([xshift=-15pt] sda.north) -- ([xshift=-15pt] sda.north |- im.south) node[edge label, midway, left, pos=0.5, align=right] {\dlfont{hasImplementation}\\\dlfont{Manifest}};
            % AssuranceSDA -> SafetyAssurance
            \draw[->, ObjectProperty] (asda.north) -- (sa.south) node[edge label, midway, right, pos=0.5, align=right] {\hSAssurancce};

            \draw[->, subClassOf] (rsda) -- (sda);
            \draw[->, subClassOf] (rsdai) -- (rsda);
            \draw[->, subClassOf] (rsdai) -- (sdai);

            \draw[->, subClassOf] (asda) -- (sda);
            \draw[->, subClassOf] (asdai) -- (asda);
            \draw[->, subClassOf] (sdai) -- (sda);
            \draw[->, subClassOf] (asdai) -- (sdai);

            \node[draw,myborder,fill=none,fit=(ev)(sev)(cr)(dcm),inner xsep=32pt,xshift=-25pt, label={[font=\scriptsize, xshift=130pt, yshift=12pt]south:\textit{Risk Assessment}}] {};

            \node[draw,myborder,fill=none,fit=(im)(asdai),inner xsep=18pt,xshift=15pt, label={[font=\scriptsize, xshift=94pt, yshift=20pt, text width=1cm]south:\textit{Risk Control}}] {};

            % Legend
            \node[left=90pt of im, draw=none, fill=none] (leg) {Legend:} ;
            \node[below=12pt of leg, minimum width=1.5cm] (sub) {subclass} ;
            \node[right=of sub, minimum width=1.5cm] (super) {superclass} ;
            \node[above=12pt of super, minimum width=1.5cm] (cls) {Class} ;
            \node[below=11pt of sub, minimum width=1.5cm] (dom) {domain} ;
            \node[right=of dom, minimum width=1.5cm] (ran) {range} ;

            \draw[->, subClassOf] (sub) -- (super) node[edge label, midway, above, align=center] {sub-\\ClassOf};
            \draw[->, ObjectProperty] (dom) -- (ran) node[edge label, midway, above, align=center] {object\\property};
      \end{tikzpicture}%
      \caption{Schema diagram of the \riskman classes and properties, divided into two sections covering the outcomes of Risk Assessment and Risk Control.
            While range restrictions have an explicit syntax, domain restrictions \mbox{$\dom{\RName}\dlsub\CName$} are expressed via \mbox{$\ex{\RName}\dlsub\CName$}, just as \mbox{$\tra{\RName}$}, saying that $\RName$ is transitive, is syntactic sugar for \mbox{$\RName\circ\RName\dlsub\RName$}.
            Moreover, any two classes without direct/indirect subclass relationship are disjoint.}
      \label{fig:schema}
\end{figure}
%
%\mysubsection{Ontology Design Patterns}
\noindent
We opted for a lightweight ontology that captures the outcomes of risk management with low ontological commitment, as the particular needs might differ from one manufacturer to another.
To ensure the usability and extensibility of the ontology, we employed the Stub Metapattern~\cite{KrisnadhiH16,StubURL}, which “acts as a type of placeholder for future extensions.”
Specifically, the class \emph{DeviceProblem} is intended to link to the \mbox{IMDRF}'s controlled vocabularies~\cite[Annex A]{IMDRF20AET};
another stub, \emph{HazardousSituation}, enables to reuse
the \emph{Hazardous Situation Pattern}~\cite{Lawrynowicz2015TheHS,Cheatham2016AMT} if a more fine-grained representation is required.\refreq{q}{2}

\mysubsection{Probabilities and Severities}
\label{sec:probabilities-severities}

\fussy
\noindent
VDE~Spec~90025 (importing from ISO~14971) defines risk as the “combination of the probability of occurrence of harm and the severity of that harm”, and so to represent concrete risks it is necessary to also represent concrete “values” for probability (and severity).
Mathematical probabilities are virtually impossible to accurately determine for events and situations that are hypothetical from the outset.
Therefore, the typical approach in risk management (cf.~\Cref{sec:background}) is to use a fixed, finite number of probability \emph{magnitudes}, each representing an \emph{interval} of real-valued probabilities, and being naturally ordered on a logarithmic scale,
e.g.\ \probmag{“improbable”}{$(0,10^{-4}]$},
\probmag{“remote”}{$(10^{-4},10^{-3}]$},
\probmag{“occasional”}{$(10^{-3}, 10^{-2}]$},
\probmag{“probable”}{$(10^{-2},10^{-1}]$},
and \probmag{“frequent”}{$(10^{-1},1)$}.
The exact number $p$ of different magnitudes varies and is up to the manufacturer;
choosing \mbox{$p=5$} (as above) is common.
A similar approach is typically also used for severity.

We refrained from binding users of the \riskman ontology to a specific way of representing probability and severity, but at the same time want to provide a reasonable baseline that can be used almost “out of the box”.
To this end, we have an optional “plugin” that creates probability and severity magnitudes (ontologically represented by individuals using nominals) for given desired interval counts $\probcount$ (probability) and $\sevcount$ (severity), together with additional axioms as an ontology \psonto{\probcount}{\sevcount} (for \emph{probability-severity ontology}),
with \mbox{$\psonto{\probcount}{\sevcount}\mathbin{:\mkern-1mu=}\psTBox\cup\psABox$} where
%Given two natural numbers $\mathit{P_{m}}$ and $\mathit{S_{m}}$, the following sets of individuals denoting probabilities and severities are generated and assigned to their respective classes.
\bgroup
\vspace*{-1ex}
\cramalign
\begin{align*}
        \psTBox & =  \left\{\,\exists \hPOne.\set{\dlprob} \sqcap \exists\hPTwo.\set{\dlprob[j]} \sqsubseteq \exists\hProbability.\set{\dlprob[k]} \right.             \\
                & \left.\qquad \vert\; 1\leq i,j\leq\probcount, k=\max(1, i+j-\probcount) \right\}\cup\set{\tra{\grt}}                                                 \\
        \psABox & =  \set{\,\cassert{\dlprob}{\Probability}\guard 1\leq i\leq\probcount}\cup\set{\cassert{\dlsev}{\Severity}\guard 1\leq i\leq\sevcount} \cup          \\
                & \qquad \set{\rassert{\dlprob[{i+1}]}{\dlprob}{\grt}\guard 1\leq i<\probcount}\cup\set{\rassert{\dlsev[{i+1}]}{\dlsev}{\grt}\guard 1\leq i<\sevcount} \\[-4ex]
\end{align*}
\egroup%
This introduces not only the discretized probability values (e.g.~for \mbox{$\probcount=5$} we  get $\dlprob[1]\represents\text{“improbable”}$ and $\dlprob[5]\represents\text{“frequent”}$), but also an ordering
%on the values that will later be used (among other things) to check that mitigation did not accidentally \emph{increase} a probability.
$\grt$ on these values.
Most importantly, however, the GCIs in $\psTBox$ implement the “computation” of overall probability $P$ from probabilities \Pone and \Ptwo~\cite{ISO14971}.
Essentially, the multiplication \mbox{$P \mathbin{:\mkern-1mu=} \Pone \cdot \Ptwo$} works by adding exponents of upper bounds of intervals, e.g.\ $(10^{-3},10^{-2}] \ast (10^{-2},10^{-1}]$ yields $(10^{-4},10^{-3}]$, and all possible computations for the given $\probcount$ are expressed via GCIs.

%%%%%%%%%%%%%%%%%%%%%%%%%%%%%%%%%%%%%%%%%%%%%%%%%%%%%%%%%%%%%%%%%%%%%%%%%%%%%%%%
\mysubsection{Shapes}

As explained above, our ontology can be used to derive implicit information via reasoning.
The materialized graph is then stored in RDF format~\cite{owlov}. %\footnote{The OWL~2 specification~\cite{owlov} requires that all tools at least support RDF/XML.}
(As an illustration, we depict the implementation of \Cref{ex:running} in \Cref{fig:abox:graph}.)
Our use case, the evaluation of risk reports, requires the capability to check for missing information  or for mismatches between values (e.g., whether a mitigation of a risk does not increase its severity or probability).
These kinds of checks on a (hopefully) complete and self-contained risk report are conceptually not a good fit with OWL's open-world assumption.
We thus
define SHACL shapes that operate on the materialized RDF graph and implement the most important requirements risk reports should fulfil.
We describe them in what follows.
%Users clearly can, and most likely will, add custom shapes.

We adopt the abstract syntax of SHACL constraints proposed by Corman et al.~\cite{CormanRS18,AndreselCORSS20}, capturing the core components of the SHACL specification~\cite{SHACL}.
For our purposes, we conveniently re-use description logic vocabulary, viz., pairwise disjoint sets $\Cons$ of \define{classes}, $\Rols$ of \define{properties}, and $\Inds$ of \define{individuals}.
A finite set $\ABox$ of assertions (an ABox) can then be seen as representing a labelled \define{graph}, with individuals acting as nodes, classes labelling nodes, and properties labelling edges.
The syntax of \define{shape expressions} $\phi$ and \define{path expressions} $E$ is shown in \Cref{fig:shacl-syntax}~(top). %the grammar
For the semantics, a given graph (ABox) \ABox with nodes (individuals) $\VGraph$ defines an evaluation function \mbox{$\geval{\cdot}$} that assigns to each path expression $E$ a binary relation \mbox{$\geval{E}\subseteq\VGraph\times\VGraph$}, and to each shape expression $\phi$ a set \mbox{$\geval{\phi}\subseteq\VGraph$} %of individuals
via induction as shown in \Cref{fig:shacl-syntax} (bottom).

\begin{table}
    \caption{Syntax and semantics of path and shape expressions.}
    \label{fig:shacl-syntax}
    \hrulefill

    The syntax of path expressions $E$ and shape expressions $\phi$ is given by the grammars
    \begin{align*}
        E \ebnfeq \RName \ebnfalt \RName^- \ebnfalt E\cup E \ebnfalt E\sbullet E \ebnfalt E^*
        \text{ and }
        \phi \ebnfeq \top \ebnfalt \CName \ebnfalt \IName \ebnfalt \phi_1\land\phi_2 \ebnfalt \neg\phi\ebnfalt\hasatleast E.\phi\ebnfalt\forall E.\phi\ebnfalt E=E
    \end{align*}%
    where $n\in\NPlus$, $\CName\in\Cons$, $\IName\in\Inds$, and $\RName\in\Rols$ with $\RName^-$ indicating the inverse of $\RName$.%
    \vspace*{4pt}
    \newcommand{\wideelem}[1]{\ensuremath{\makebox[3cm][l]{$#1$}}}
    \newcommand{\skipbelowline}{\raisebox{2pt}[3ex][0pt]{}}
    \bgroup
    \cramalign
    \begin{align*}
        \hline
        \skipbelowline
        \geval{\RName}                    & =\set{\tuple{a,b}\guard \RName(a,b) \in \Graph}
                                          &
        \geval{E_1\sbullet E_2}           & =\geval{E_1}\circ\geval{E_2}                                                                                                                                                             \\
        \geval{\RName^-}                  & =\set{\tuple{b,a}\guard \RName(a,b) \in \Graph}
                                          &
        \hspace*{-1em}\geval{E_1\cup E_2} & =\geval{E_1}\cup\geval{E_2}
                                          &
        \hspace*{-1ex}\geval{E^*}         & = \left(\geval{E}\right)^*                                                                                                                                                               %=\textstyle\bigcup_{n\geq 0}\left(\geval{E}\right)^n                                                                                                                             \\[2pt]
        \\
        %\hline
        \geval{\top}                      & =\VGraph                                                                                           & \geval{\CName} & =\set{a\guard \CName(a)\in\Graph} & \geval{\IName} & =\set{\IName} \\
        \geval{\phi_1\land\phi_2}         & =\geval{\phi_1}\cap\geval{\phi_2}                                                                  &
        \geval{\neg \phi}                 & =\VGraph\setminus\geval{\phi}                                                                                                                                                            \\
        \geval{\forall E.\phi}            & = \wideelem{\set{a\guard\forall b:\tuple{a,b}\in\geval{E}\text{ implies }b\in\geval{\phi}}}                                                                                              \\
        \geval{\hasatleast[n] E.\phi}     & = \wideelem{\set{a\guard\card{\set{\tuple{a,b}\in\geval{E}\text{ and }b\in\geval{\phi}}} \geq n}}                                                                                        \\
        \geval{E_1=E_2}                   & = \wideelem{\set{a\guard\forall b:\tuple{a,b}\in\geval{E_1}\text{ iff }\tuple{a,b}\in\geval{E_2}}}                                                                                       \\[2pt]
        \hline
    \end{align*}%
    \vspace*{-4ex}
    \egroup
\end{table}

A \define{shape constraint} is an expression of the form $\CName\leftarrow\phi$, with $\CName\in\Cons$ and $\phi$ a shape expression.
A \define{shape schema} is a pair $\tuple{\Constraints,\Targets}$ where $\Constraints$ is a set of shape constraints and $\Targets$ is a set of \define{target} concept assertions.
Intuitively, a target $\cassert{\IName}{\CName}$ expresses the requirement that $\IName$ be labelled by $\CName$.
Formally, an ABox $\ABox$ is a \define{model} for a set $\Constraints$ of constraints if $\geval{\phi}\subseteq\geval{\CName}$ for all $\CName\gets\phi\in\Constraints$.
%A set $\Targets'$ of concept assertions \define{validates} an ABox $\ABox$ against a schema $\tuple{\Constraints,\Targets}$ iff
An ABox $\ABox$ is \define{validated} against a schema $\tuple{\Constraints,\Targets}$ if there exists a set $\Targets'$ of concept assertions such that
(1) $\Targets\subseteq\Targets'$,
(2) $\Inds(\Targets')\subseteq\Inds(\ABox)$, and
(3) $\ABox\cup\Targets'$ is a model for $\Constraints$.
The shapes constraints that complement the \riskman ontology are listed in \Cref{fig:shacl-constraints}.
To illustrate their use, in \Cref{fig:abox:graph} individuals satisfying the shape expression part of a constraint are labelled by the constraint's number.
\begin{figure}
    \smaller
    \bgroup
    \cramalign
    \begin{align}
        \begin{split}\label{shape:arisk}
            \ARisk & \leftarrow  \hasexactlytop{\hDSH}\land\hasexactlytop{\hHarm} \land \hasexactlytop{\hDContext} \\
                   & \phantom{\leftarrow}\quad\land \hasexactlytop{\hIRL} \land \hasexactlytop{\hHSituation}
        \end{split}                    \\
        \begin{split}\label{shape:asda}
            \ASDA & \leftarrow \forall\hSSDA.\ASDA \land \hasexactlytop{\hSAssurancce}
        \end{split}                                                \\
        \begin{split}\label{shape:crisk:one}
            \CRisk & \leftarrow \hasexactlytop{\iMBy}\land\hasexactlytop{\hARisk}\land\hasexactlytop{\hRRL}
        \end{split}                           \\
        \begin{split}\label{shape:crisk:two}
            \CRisk & \leftarrow  \hARisk\sbullet\hIRL\sbullet\dlfont{X}\sbullet\grt^-\sbullet\dlfont{X}^- \neq\hRRL
        \end{split}                   \\
        \begin{split}\label{shape:dsh}
            \hspace*{-4em}\DSH & \leftarrow \hasexactlytop{\hDComponent}\land\hasexactlytop{\hDFunction}\land\hasexactlytop{\hHazard}
        \end{split} \\
        \RLevel            & \leftarrow  \hasexactlytop{\hProbability}\land\hasexactlytop{\hSeverity}\label{shape:rlevel}         \\
        \hspace*{-2em}\SDA & \leftarrow  \exists\hSSDA^*.\SDAI\label{shape:sda}
    \end{align}%
    \egroup
    % }%
    \caption{\riskman shape constraints. The following syntactic abbreviations are used for brevity:  $\exists E.\phi$ for $\hasatleast[1] E.\phi$, $\hasatmost E.\phi$ for $\neg(\hasatleast[n+1] E.\phi)$, $\hasexactly[1] E.\phi$ for $\hasatleast[1] E.\phi\land\hasatmost[1] E.\phi$, $E\neq E'$ for $\neg(E=E')$. For Constraint~\ref{shape:crisk:two} we denote \mbox{$\dlfont{X}\in\set{\hProbability, \hPOne, \hPTwo, \hSeverity}$}.}
    \label{fig:shacl-constraints}
\end{figure}

Constraint~\ref{shape:crisk:two} encodes checks for non-increasing residual risk levels, i.e. whether the probability or severity after implementing a mitigation is not higher than before.
Constraints~\ref{shape:arisk},~\ref{shape:crisk:one},~\ref{shape:dsh},~and~\ref{shape:rlevel} are duals of the subclass declarations for \ARisk, \CRisk, \DSH, and \RLevel, respectively, in that they enforce instances of these classes to contain all the necessary components of the class definitions, as required by \vdespec.
% ({In case of \DSH, in contrast to Axiom~\ref{gci:dsh:two}, existence of a \DProblem\xspace is only an optional way of annotating a \CRisk\xspace with an IMDRF code, and as such is not required by~\ref{shape:dsh}.}) % Not required since we removed DeviceProblem from the definition.
Constraint~\ref{shape:asda} encodes that (i) every sub-\dlfont{SDA}\xspace of an \ASDA\xspace must be an \ASDA\xspace and (ii) an \ASDA\xspace must have a \SAssurancce.
Assuming the ABox has previously been materialized not only by means of the \riskman ontology, but also the additional ``probability-severity'' ontology \psonto{\probcount}{\sevcount}, Constraint~\ref{shape:rlevel} requires that
(i) an overall probability $P$ is present (specified directly or inferred from probabilities $\Pone$ and $\Ptwo$ via \psonto{\probcount}{\sevcount}),
(ii) that \mbox{$P=\Pone\cdot\Ptwo$} holds in case all three have been specified, and
(iii) a severity magnitude is present.
Finally, the check whether all leaf nodes of the SDA~tree are \SDAI s is provided by Constraint~\ref{shape:sda}.
\Cref{fig:abox:graph} illustrates how to use our ontology and shapes by implementing the example of \Cref{fig:rmf}.

% ABox
\begin{figure}[h]
    \centering
    % \vskip4pt
    \bgroup
    \hypersetup{linkcolor=red}
    \begin{tikzpicture}
        \tikzset {
            ind/.style={
                    draw=black!5,
                    rectangle,
                    rounded corners,
                    text centered,
                    font=\scriptsize,
                    minimum width=15pt,
                    minimum height=15pt,
                    inner sep=0pt,
                    fill=black!5,
                },
            edg/.style={
                    sloped, above,
                    midway, font=\tiny
                },
            infedg/.style={
                    blue, dashed
                },
            inf/.style={
                    font=\tiny,
                    text=blue,
                    align=left
                },
            shp/.style={
                    font=\tiny,
                    text=red,
                },
        }
        % ControlledRisk
        \node[ind] (ctr1) {\ind{cr}};
        \node[] (infctr1) [below=3pt of ctr1, inf, align=center] {\Risk,\\\CRisk} ;
        \node[] (shapectr1) [above right=0pt and 0pt of ctr1, xshift=-10pt, yshift=-4pt, shp, align=center] {(\ref{shape:crisk:one},\ref{shape:crisk:two})} ;

        % SDA
        \node[ind] (sda0) [left=2cm of ctr1] {\indid{sd}{0}};
        \node[] (infsda0) [above=0pt of sda0, inf] {\SDAShort} ;
        \node[] (shapesda0) [above=1pt of sda0, yshift=5pt, shp] {(\ref{shape:sda})} ;

        \node[ind] (sda2) [below=of sda0] {\indid{sd}{2}};
        \node[] (infsda2) [below=0pt of sda2, inf] {\SDAShort,\\\SDAI} ;
        \node[] (shapesda2) [right=0pt of sda2, shp, xshift=-2pt] {(\ref{shape:sda})} ;

        \node[ind] (sda1) [left=of sda2] {\indid{sd}{1}};
        \node[] (infsda1) [above left=0pt and 0pt of sda1, xshift=15pt, inf] {\SDAShort,\\\SDAI} ;
        \node[] (shapesda1) [above left=0pt and 0pt of sda1, shp, yshift=-10pt] {(\ref{shape:sda})} ;

        \node[ind] (sda3) [right=of sda2] {\indid{sd}{3}};
        \node[] (infsda3) [right=0pt of sda3, inf, xshift=-2pt] {\SDAShort} ;
        \node[] (shapesda3) [right=0pt of sda3, shp, yshift=7pt, xshift=-2pt] {(\ref{shape:sda})} ;

        \node[ind] (sda4) [below=1.25cm of sda3] {\indid{sd}{4}};
        \node[] (infsda4) [left=0pt of sda4, yshift=3pt, inf] {\SDAShort,\\\SDAI} ;
        \node[] (shapesda4) [above left=0pt and 0pt of sda4, shp, align=center] {(\ref{shape:sda})} ;

        \node[ind] (sda5) [right=1.25cm of sda4] {\indid{sd}{5}};
        \node[] (infsda5) [above right=1pt and 0pt of sda5, xshift=-15pt, inf] {\SDAShort,\SDAI\\ \ASDA,\\ \ASDAI} ;
        \node[] (shapesda5) [below=0pt and 0pt of sda5, shp, align=center] {(\ref{shape:sda},\ref{shape:asda})} ;

        \node[ind] (sa) [right=1.25cm of sda5] {\ind{sa}};
        \node[] (infsa) [below=1pt of sa, inf] {\SAssurancce} ;

        % ImplementationManifest
        \node[ind] (im1) [below left=1.2cm of sda1] {\indid{im}{1}};
        \node[] (infim1) [below=1pt of im1, inf] {\IManifestShort} ;

        \node[ind] (im2) [below left=1.2cm of sda2] {\indid{im}{2}};
        \node[] (infim2) [below=1pt of im2, inf] {\IManifestShort} ;

        \node[ind] (im4) [below left=1.2cm of sda4] {\indid{im}{4}};
        \node[] (infim4) [below=1pt of im4, inf] {\IManifestShort} ;

        \node[ind] (im5) [below left=1.2cm of sda5] {\indid{im}{5}};
        \node[] (infim5) [below=1pt of im5, inf] {\IManifestShort} ;

        % ResidualRiskLevel
        \node[ind] (rrl) [below right=1.5cm and 1.5cm of ctr1] {\ind{rrl}};
        \node[] (infrrl) [right=1pt of rrl, inf] {\RLevel} ;
        \node[] (shaperrl) [below right=0pt and 0pt of rrl, yshift=5pt, shp] {(\ref{shape:rlevel})} ;

        \node[ind] (rprob) [above right=of rrl] {\indid{p}{3}};
        \node[] (infrprob) [right=1pt of rprob, inf] {\Probability} ;

        % AnalysedRisk
        \node[ind] (ar1) [above=1.5cm of ctr1] {\ind{ar}};
        \node[] (infind) [right=1pt of ar1, inf] {\Risk, \ARisk} ;
        \node[] (shapear1) [right=0pt of ar1, yshift=6pt, shp] {(\ref{shape:arisk})} ;

        % InitialRiskLevel
        \node[ind] (irl) [above right=1.5cm and 1.5cm of ar1] {\ind{irl}};
        \node[] (infirl) [above left=-3pt and -10pt of irl, inf] {\RLevel} ;
        \node[] (shapeirl) [below=0pt and 0pt of irl, yshift=-7pt, shp] {(\ref{shape:rlevel})} ;

        \node[ind] (ip2) [right=1.45cm of irl] {\indid{p}{4}};
        \node[] (infip2) [right=0pt of ip2, inf, xshift=-7pt,yshift=-15pt] {\Probability} ;

        \node[ind] (ip1) [above right=1cm and 1cm of irl] {\indid{p}{5}};
        \node[] (infip1) [above=1pt of ip1, inf] {\Probability} ;

        \node[ind] (isev) [below right=1.25cm and 1.25 of irl] {\indid{s}{4}};
        \node[] (infisev) [above=1pt of isev, inf] {\Severity} ;

        % HazardousSituation
        \node[ind] (hazsit) [above=2.5cm of ar1] {\ind{hs}};
        \node[] (infhazsit) [above left=1pt and -10pt of hazsit, inf] {\HSituation} ;

        % Event 2
        \node[ind] (ev2) [right=1cm of hazsit] {\indid{ev}{2}};
        \node[] (infev2) [above=-3pt of ev2, xshift=10pt, inf] {\Event} ;

        % Event 1
        \node[ind] (ev1) [above left=1.5cm of ev2] {\indid{ev}{1}};
        \node[] (infev1) [left=1pt of ev1, inf] {\Event} ;

        % DeviceContext
        \node[ind] (dcon) [below right=of ar1] {\ind{dcx}};
        \node[] (infdcon) [below=1pt of dcon, inf] {\DContext} ;

        % Harm
        \node[ind] (harm) [below left=.5cm and 1cm of ar1] {\ind{hr}};
        \node[] (infharm) [above=1pt of harm, inf] {\Harm} ;

        % Patient problem
        \node[ind] (pp) [left=1.7cm of ar1] {\ind{pp}};
        \node[] (infpp) [above=1pt of pp, inf] {\PProblem} ;

        % DeviceSpecificHazard
        \node[ind] (dsh) [above left=1.5cm and 1.5cm of ar1] {\ind{dsh}};
        \node[] (infdsh) [above right=1pt and 1pt of dsh, xshift=-11pt, yshift=-3pt, inf] {\DSHShort} ;
        \node[] (shapedsh) [above right=0pt and 0pt of dsh, xshift=-9pt, yshift=5pt, shp] {(\ref{shape:dsh})} ;

        % Hazard
        \node[ind] (haz) [above=1.5cm of dsh] {\ind{hz}};
        \node[] (infhaz) [above=1pt of haz, inf] {\Hazard} ;

        % DeviceComponent
        \node[ind] (dcom) [above left=1.25cm and 1.25cm of dsh] {\ind{dcm}};
        \node[] (infdcom) [above=1pt of dcom, inf] {\DComponent} ;

        % DeviceFunction
        \node[ind] (df) [left=1.75cm of dsh] {\ind{df}};
        \node[] (infdf) [above=1pt of df, inf] {\DFunction} ;

        % DeviceProblem
        \node[ind] (dp) [below left=1.25cm and 1.25cm of dsh] {\ind{dp}};
        \node[] (infdp) [above left=1pt and -10pt of dp, inf] {\DProblem} ;

        % Inferred edges
        % ControlledRisk -> Harm
        \path[->] (ctr1)  edge[bend right=10, infedg, edg, pos=0.6] node {\textcolor{blue}{\hHarm}} (harm);

        % Asserted edges
        % SDA
        \draw[->] (ctr1) -- node[edg] {\iMBy} (sda0);
        \draw[->, edg] (sda0) -- node {\hSSDA} (sda1);
        \draw[->, edg] (sda0) -- node {\hSSDA} (sda2);
        \draw[->, edg] (sda0) -- node {\hSSDA} (sda3);

        \draw[->, edg] (sda1) -- node {\hIManifestShort} (im1);
        \draw[->, edg] (sda2) -- node {\hIManifestShort} (im2);
        \draw[->, edg] (sda3) -- node {\hSSDA} (sda4);
        \draw[->, edg] (sda3) -- node {\hSSDA} (sda5);
        \draw[->, edg] (sda5) -- node {\hSAssurancceShort} (sa);

        \draw[->, edg] (sda4) -- node {\hIManifestShort} (im4);
        \draw[->, edg] (sda5) -- node {\hIManifestShort} (im5);

        % ResidualRiskLevel
        % Asserted & Inferred
        \draw[->, align=center] (ctr1) -- node[edg] {\textcolor{blue}{\hRL},\\\hRRL} (rrl);
        \draw[->] (rrl) -- node[edg] {\hProbability} (rprob);
        \path[->] (rrl)  edge[bend left=15] node[midway, above, sloped, font=\tiny, xshift=15pt] {\hSeverity} (isev);

        % AR
        \draw[->] (ctr1) -- node[edg] {\hARisk} (ar1);
        \draw[->,align=center] (ar1) -- node[edg] {\textcolor{blue}{\hRL},\\\hIRL} (irl);
        % Inferred hasProbability
        \path[->] (irl)  edge[bend left=20, infedg, edg] node[xshift=5pt,yshift=-3pt] {\textcolor{blue}{\hProbability}} (ip2);

        % Initial Risk Level
        \draw[->] (irl) -- node[edg, below] {\hPTwo} (ip2);
        \draw[->] (irl) -- node[edg] {\hPOne} (ip1);
        \draw[->] (irl) -- node[edg, below] {\hSeverity} (isev);

        % Inferred GT relations
        \path[->] (ip1)  edge[bend left=15, infedg, edg] node {\textcolor{blue}{\grt}} (ip2);
        \path[->] (ip2)  edge[bend left=15, infedg, edg] node[rotate=180] {\textcolor{blue}{\grt}} (rprob);
        \path[->] (ip1)  edge[bend left=60, infedg, edg] node {\textcolor{blue}{\grt}} (rprob);

        % AnalysedRisk
        \draw[->] (ar1) -- node[edg] {\hDSHShort} (dsh);
        \draw[->] (ar1) -- node[edg, xshift=5pt] {\hHSituation} (hazsit);
        \draw[->] (ar1) -- node[edg] {\hDContext} (dcon);
        \draw[->] (ar1) -- node[edg] {\hPProblem} (pp);
        \draw[->] (ar1) -- node[edg] {\hHarm} (harm);

        % HazardousSituation
        \draw[->] (ev2) -- node[edg] {\hPEvent} (ev1);
        \draw[->] (hazsit) -- node[edg] {\hEvent} (ev2);

        % DeviceSpecificHazard
        \draw[->] (dsh) -- node[edg] {\hDProblem} (dp);
        \draw[->] (dsh) -- node[edg] {\hDFunction} (df);
        \draw[->] (dsh) -- node[edg] {\hDComponent} (dcom);
        \draw[->] (dsh) -- node[edg, pos=0.6] {\hHazard} (haz);
    \end{tikzpicture}
    \egroup
    \caption{Graphical ABox representation of data from \Cref{fig:rmf}.
        Nodes and edges represent domain elements and role assertions, respectively.
        Correspondence between respective elements of \Cref{fig:rmf} and nodes can be established by their identifiers, with node \ind{cr} (\CRisk) being the central entry point of the graph.
        Probability and severity nodes (\indid{p}{5}, \indid{p}{4}, \indid{p}{3}, and \indid{s}{4}) correspond to individual names from \psonto{5}{5} and are interpreted by themselves.
        Black colour represents the asserted, whereas blue the inferred knowledge, involving either classes (labels near nodes) or roles (labels above edges or additional dashed edges).
        Given a constraint of the form $\CName\leftarrow\phi$ from \Cref{fig:shacl-constraints} labelled by some number $(n)$, the same number $n$ in red next to a node indicates that the node satisfies $\phi$.
        Putting all the above together, note e.g., that labels \dlfont{SDA} and \SDAI\xspace of \indid{sd}{1} indicate that it has been classified as \dlfont{SDA} due to being a \hSSDA-successor (range restriction) and as \SDAI, due to the previous classification and existence of an \hIManifest-successor (Axiom~\ref{gci:sdai}).
        On the other hand, note that \ind{irl} gained \indid{p}{4} as its \hProbability-successor due to an inference using a “multiplication” axiom from \psonto{5}{5}.
        It hence contains exactly one \hProbability\xspace and \hSeverity\xspace successor each and, therefore, satisfies the body of Constraint~\ref{shape:rlevel}, as indicated with the (\ref{shape:rlevel}) in red.
        This holds for every node labelled with \RLevel, and therefore the depicted ABox satisfies Constraint~\ref{shape:rlevel}.}
    \label{fig:abox:graph}
\end{figure}

%%%%%%%%%%%%%%%%%%%%%%%%%%%%%%%%%%%%%%%%%%%%%%%%%%%%%%%%%%%%%%%%%%%%%%%%%%%%%%%%
\mysubsection{Extensibility}

Alas, even a carefully designed framework like ours, which takes the needs of the different stakeholders into account, can never meet everyone's wishes.
Companies could have individual requirements, authorities might want to add specialized tests, and regulations differ per region and change over time.
%This is one of the main reasons why we chose Semantic Web technologies for our task: they are easy to extend and can be combined with existing data.
To illustrate the extensibility of our approach, we give an example:
We described in \Cref{sec:probabilities-severities} that companies often define magnitude levels to model probabilities and severities of risks.
%While studying risk reports, we observed that 
In practice,
these definitions sometimes come with a so-called \emph{risk acceptance matrix}, a schema indicating which combinations of probability and severity a company considers critical.
This can be modelled with our ontology by adding a class \CRL{}.
Assume now for the sake of example that combination “probability $\dlprob[5]$ and severity $\dlsev[3]$” is critical.
This can be stated via an axiom
\mbox{$\exists\hProbability.\{\dlprob[5]\}\sqcap\exists\hSeverity.\{\dlsev[3]\}\sqsubseteq \CRL$}
allowing to easily check for controlled risks with critical residual risk levels using SHACL via a constraint
\mbox{$\CRisk \leftarrow \neg(\exists\hRRL.\CRL)$}.
Of course, more complex additions of further shapes and constraints are likewise possible.

%%%%%%%%%%%%%%%%%%%%%%%%%%%%%%%%%%%%%%%%%%%%%%%%%%%%%%%%%%%%%%%%%%%%%%%%%%%%%%%%
\FloatBarrier
\mysection{Discussion and Outlook}
\label{sec:discussion}

%% conclusion

We presented the \riskman ontology and shapes with their intended use of representing and analysing risk management information for medical devices.
Analyzed risks and their mitigations are represented as an \ELpp~ABox, the \riskman ontology is used with a reasoner to infer implicit knowledge, and lastly SHACL constraints are used to check whether the input data conform to given requirements.
The ontology and shapes constraints are freely \href{https://w3id.org/riskman}{available}, including a reference implementation of the whole pipeline.
With feedback from manufacturers and notified bodies incorporated into it, we envision \riskman to improve the work lives of risk managers and certification auditors alike.\refreq{r}{1}

%% future work

The issue of medical device safety will only become more important in the future, especially with further digitization~\cite{SujanSC19}.
This is the case even more so for devices that use artificial intelligence themselves, and as such fall under the regulations of the \emph{AI Act}~\cite{AIAct};
an integration with the like-minded AIRO ontology~\cite{GolpayeganiPL22} is an important topic for future work.
An interesting next step for the \riskman ontology is to not only assess submission completeness, but also evaluate the quality of the assurance~\cite{ChowdhuryWPL20}.
First steps, including a proof of concept, have recently been achieved in the context of assurance cases~\cite{FosterNGWK21}, but are known to be hard to generalize~\cite{delaVaraJMP19}.
Possible techniques to consider for evaluation of risk management artifacts represented using \riskman are dialogue-based approaches to proof theory in structured argumentation~\cite{DillerGG21}, as well as other approaches to provide justifications in logic-based knowledge representation formalisms~\cite{DeneckerBS15}.

%In the future, we also want to use existing data about adverse events (partially annotated with terms from IMDRF taxonomies, e.g.\ from the FDA's Manufacturer and User Device Facility Experience database, \href{https://www.accessdata.fda.gov/scripts/cdrh/cfdocs/cfmaude/Search.cfm}{MAUDE}~\cite{MAUDE}%\footnote{\url{https://www.accessdata.fda.gov/scripts/cdrh/cfdocs/cfmaude/Search.cfm}.}) to allow for a more informed evaluation of “whether all hazards have been identified“, a classical point of risk management that cannot strictly be proved by a manufacturer, but potentially be disproved by a sufficiently knowledgeable reviewer/auditor.

Ahmetaj et al.~\cite{AhmetajDOPSS21} analyzed how non-validation of SHACL constraints can be explained to users (in terms of \emph{repairs}), which can potentially be applied to our work and be included in a future \riskman-based work bench for risk managers.

Instead of delegating the inference and validation steps to two different services (reasoners/validators), we could also utilize advancements in the area of combining OWL and SHACL, e.g.\ by converting the ontology and shapes into a single set of SHACL constraints~\cite{AhmetajOOS23}.
While this involves an exponential blowup in general~\cite{AhmetajOOS23}, the fact that \riskman stays within \ELpp might constitute an interesting special case.
Alternatively, having a single reasoner perform inferencing as well as constraint checking could also be achieved by translating ontology axioms and shapes constraints into answer set programming~\cite{BrewkaET11}.
Different implementations could then be compared experimentally.

As a possible alternative to extending some concepts that are currently stubs (in the sense of the stub metapattern~\cite{KrisnadhiH16}), we envision to use the novel formalism of \emph{standpoint logic}~\cite{AlvarezR21,AlvarezRS22} to import and attach further ontologies to \riskman.\refreq{r}{4}
In this regard it is especially notable and useful that the combination of standpoint logic and the description logic \EL retains the latter's polynomial time computational complexity~\cite{AlvarezRS23a,AlvarezRS23b}.
A prominent candidate for integration is the US National Cancer Institute's thesaurus (NCIt)~\cite{ncitURL}, %\footnote{Available for browsing and download at \url{http://ncit.nci.nih.gov/}.}
which can be expressed in \ELpp~\cite{BaaderLB08}.
On the other hand, \riskman itself could also be embedded into a top-level ontology, e.g.\ BFO~\cite{Arp15} or GFO~\cite{Herre10}.

\FloatBarrier
% \input{sections/resource-availability-statement}

% %%%%%%%%%%%%%%%%%%%%%%%%%%%%%%%%%%%%%%%%%%%%%%%%%%%%%%%%%%%%%%%%%%%%%%%%%%%%%%%%
\newcommand{\projectname}[1]{\mbox{#1}}
\subsubsection*{Acknowledgements.}
This work was supported by funding from BMFTR (Federal Ministry of Research, Technology and Space) within projects
\projectname{KIMEDS} (grant no.~GW0552B),
\projectname{MEDGE} (grant no.~16ME0529),
\projectname{SEMECO} (grant no.~03ZU1210B), and
\projectname{SECAI} (within DAAD project 57616814, School of Embedded Composite AI, \url{https://secai.org/}, as part of the program Konrad Zuse Schools of Excellence in Artificial Intelligence).

We are indebted to numerous people who explained the theory and practice of risk management, commented on our earlier formalisations, and participated in discussions that improved our understanding of the domain.
In alphabetical order, we thank:
Philipp Bank,
Manuel Baur,
Samet Bayraktar,
Gustav Bieberstein,
Nidhal Chouchane,
Tony Dietrich,
Anne Eßlinger,
Ludger Evers,
Felix Gebhardt,
Stephen Gilbert,
Andreas Halbleib,
Christian Helmbold,
Evi Hartig,
Aleksandr Ilinykh,
Abtin Jamshidirad,
Simon Kilcher,
Jürgen Koch,
Tina Küttner,
Andreas Lämmerzahl,
Felix Lempke,
Alf Ludwig,
Bokhodir Mamadaliev,
Svetlana Miasoedova,
Davood Moghadas,
Martin Neumann,
Andreas Purde,
Katharina Rehde,
Christian Rosenzweig,
Kerstin Rothweiler,
Jasmine Schirmer,
Hubert Sefkovicz,
Sebastian Schostek,
Robin Seidel,
Robert Stelzmann,
Sarah Tsurkan,
Heike Vocke,
Hans Wenner,
Martin Witte,
Sven Wittorf,
Juliane Wober,
and
Uwe Zeller.


\bibliographystyle{vancouver}
\bibliography{bib/references}

\iflong

      \clearpage

      \appendix

      \section{Definitions of Terms from VDE~Spec~90025}
      \label{sec:app:defs}
      In this section, we briefly recall relevant definitions from VDE~Spec~90025~\cite{VDESpec24}.     \Cref{tab:terms_iso} provides definitions of terms borrowed from \isonormcite, while    \cref{tab:terms_vde} gives definitions of terms newly introduced in VDE~Spec~90025.

\begin{table}[h]
    \centering
    \caption{ISO 14971 terms and definitions}
    \label{tab:terms_iso}
    \begin{tabular}{p{2cm} |  p{9cm}}
        %\hline
        \textbf{Term}                  & \textbf{Definition}                                                                                                                                             \\
        \hline
        Harm                           & \emph{Injury or damage to the health of people, or damage to property or the environment.}                                                                      \\
        \hline
        Hazard                         & \textit{Potential source of harm.}                                                                                                                              \\
        \hline
        Hazardous situation            & \textit{Circumstance in which people, property, or the environment is/are exposed to one or more hazards.}                                                      \\
        \hline
        Intended use, intended purpose & \textit{Use for which a product, process, or service is intended according to the specifications, instructions140and information provided by the manufacturer.} \\
        \hline
        Objective evidence             & \textit{Data supporting the existence of verity of something.}                                                                                                  \\
        \hline
        \Pone                          & \textit{Probability of the occurrence of a hazardous situation.}                                                                                                \\
        \hline
        \Ptwo                          & \textit{Probability of a hazardous situation leading to harm.}                                                                                                  \\
        \hline
        Residual risk                  & \textit{Risk remaining after risk control measures have been implemented.}                                                                                      \\
        \hline
        Risk                           & \textit{Combination of the probability of occurrence of harm and the severity of that harm.}                                                                    \\
        \hline
        Risk analysis                  & \textit{Systematic use of available information to identify hazards and to estimate the risk.}                                                                  \\
        \hline
        Risk control                   & \textit{Process in which decisions are made and measures implemented by which risks are reduced to, or maintained within, specified levels.}                    \\
        \hline
        Safety                         & \textit{Freedom from unacceptable risk.}                                                                                                                        \\
        \hline
        Severity                       & \textit{Measure of the possible consequences of a hazard.}                                                                                                      \\
        \hline
        State of the art               & \textit{Developed stage of technical capability at a given time as regards products, processes, and services,
        based on the relevant consolidated findings of science, technology, and experience.}                                                                                                             \\
        \hline
        %\hline
    \end{tabular}
\end{table}

\begin{table}[h]
    \centering
    \caption{VDE Spec 90025 new terms and definitions}
    \label{tab:terms_vde}
    \begin{tabular}{p{2.5cm} |  p{8cm}}
        %\hline
        \textbf{Term}               & \textbf{Definition}                                                                                                                                                                                \\
        \hline
        Analyzed risk               & \textit{Combination of one or more domain-specific hazard(s) with one hazardous situation and one harm with reference to a device context and a specification of an initial risk level.}           \\
        \hline
        Assurance SDA               & \textit{SDA where the purpose is to make a safety assurance.}                                                                                                                                      \\
        \hline
        Assurance SDAI              & \textit{SDAI of an assurance SDA.}                                                                                                                                                                 \\
        \hline
        Controlled risk             & \textit{Structured artifact that relates one analyzed risk with one or more SDA(s) and specifies a resulting residual risk.}                                                                       \\
        \hline
        Device component            & \textit{A (physical or logical) part of a device.}                                                                                                                                                 \\
        \hline
        Device context              & \textit{Information concerning the use context of a device, including, but not limited to, (1) intended use/intended purpose, (2) instructions for use, and (3) intended environment of use.}      \\
        \hline
        Device function             & \textit{Functional device capability at application level.}                                                                                                                                        \\
        \hline
        Domain-specific hazard      & \textit{Structured artifact that centres around one hazard having the potential to cause one or more harm(s) in the context of a domain-specific function and component.}                          \\
        \hline
        Event                       & \textit{Atomic occurrence or incident that (possibly when linked in a sequence with other events) may spawn a hazardous situation from a domain-specific hazard.}                                  \\
        \hline
        Implementation manifest     & \textit{Concrete piece of objective evidence (or a reference to such) that an SDA has been implemented, e.g. reference to a line of code or a particular section in the device manual.}            \\
        \hline
        Intended environment of use & \textit{Environment or environmental conditions in which the device is intended to be used.}                                                                                                       \\
        \hline
        Risk matrix                 & \textit{Matrix (two-dimensional table) displaying all combinations of probability and severity classes without determining which of those combinations are acceptable.}                            \\
        \hline
        Risk SDA                    & \textit{SDA where the purpose is to control a Risk.}                                                                                                                                               \\
        \hline
        Risk SDAI                   & \textit{SDAI of a Risk SDA.}                                                                                                                                                                       \\
        \hline
        Risk level                  & \textit{Combination of probability and severity.}                                                                                                                                                  \\
        \hline
        Safety assurance            & \textit{A credible reference (or list of such) to the state of the art of achieving safety with respect to a certain class of hazards, e.g. referring to an international norm such as IEC 60601.} \\
        \hline
        SDA (Safe design argument)  & \textit{Reusable artifact embodying or expressing one possible method or approach towards a specific goal.}                                                                                        \\
        \hline
        SDAI (SDA implementation)   & \textit{Structured artifact specifying a concrete implementation or realisation of a specific SDA.}                                                                                                \\
        \hline
        Use-Context                 & \textit{Intended/reasonably foreseeable environment the device can be used in, that may affect a related risk.}                                                                                    \\
        \hline
        %\hline
    \end{tabular}
\end{table}

\fi

% \section{Requirements}
% \input{sections/requirements}

\end{document}